\documentclass{article} 
\usepackage{iclr2027_conference,times}

\usepackage{amsmath,amsfonts,bm}

\def\eqref#1{equation~\ref{#1}}

\def\1{\bm{1}}

\DeclareMathAlphabet{\mathsfit}{\encodingdefault}{\sfdefault}{m}{sl}
\SetMathAlphabet{\mathsfit}{bold}{\encodingdefault}{\sfdefault}{bx}{n}

\usepackage{hyperref}
\usepackage{url}
\usepackage{graphicx}
\usepackage{amsmath}
\usepackage{amssymb}
\usepackage{booktabs}
\usepackage{multirow}
\usepackage{adjustbox}
\usepackage{wrapfig}
\usepackage[font=small]{caption}
\usepackage{xcolor}

\title{BeatGraph: Self-Supervised Heartbeat Graphs for Infant ECG Representations from the Home Environment}

\author{%
Mohammad Nur Hossain Khan$^{1}$, M.~S.~Krafczyk$^{2}$, Beverly G.~Bolster$^{2}$, \\
\textbf{Nancy McElwain$^{2}$, Mark A.~Hasegawa-Johnson$^{2}$, Bashima Islam$^{1}$} \\
$^{1}$University of Massachusetts Amherst \quad
$^{2}$University of Illinois Urbana-Champaign \\
}

\iclrfinalcopy 
\begin{document}

\maketitle

\begin{abstract}
Electrocardiogram (ECG) foundation models typically tokenize the signal into fixed-length patches that ignore cardiac structure, so a patch may split a heartbeat and the number of beats contained in each patch shifts as heart rate varies. This matters most for infants, whose heart rates are higher than adults and whose ECG differs from the adult, clinic-recorded 12-lead data on which these models are built. A model for infant ECG should therefore reason about heartbeats directly rather than recover them from arbitrary patches. We propose BeatGraph, which makes the heartbeat its unit of representation, modeling each 30-second window as a graph of beats. A shared beat encoder turns each heartbeat into a feature vector from its waveform and inter-beat intervals, a Transformer with positional encoding orders the beats in time, and residual graph attention layers relate every beat to every other before attention-based pooling produces a window-level representation. We pretrain BeatGraph on our new corpus of unlabeled infant recordings by predicting the embeddings of masked beats, then fine-tune it for each task. This single backbone supports sleep-wake detection, infant-state classification, activity-source identification, meaning whether movement is infant- or caregiver-initiated, and infant affect recognition, improving macro-F1 over the strongest baseline on each task by 0.076 to 0.158. It also transfers across age groups, reaching 0.892 AUROC on the ZZU-pECG pediatric benchmark among children 0 to 14 years of age, within 0.001 of the best published self-supervised ECG model, and matching that model under linear evaluation on the adult PTB-XL benchmark despite pretraining only on infant recordings. Finally, to our knowledge, we release the first public infant ECG corpus collected in homes, classrooms, and laboratory settings with state and affect labels. This corpus contains 3,408 hours of single-channel ECG from 143 infants aged 3 to 11 months, with unlabeled pretraining data, benchmark tasks, and subject-level splits.
\end{abstract}

\section{Introduction}

\label{introduction}

Beat-to-beat cardiac activity is regulated by the autonomic nervous system, which shows immense development during the first years of life ~\citep{porges2011,harteveld2021}. This cardiac activity tracks sleep and wake organization, arousal, attention, and emotion regulation, and heart-rate variability, in particular, is a well established marker of infant autonomic regulation that undergirds such behavioral states ~\citep{porges2011,graziano2013}. Audio and video recordings can capture observable behaviors, yet a quietly distressed or drowsy infant leaves little trace in either. Because electrocardiography (ECG) indexes autonomic function directly, it can be used to assess a wider array of infant states, including those with no outward behavioral expression. Useful representations of infant ECG could therefore support unobtrusive measurements of infants' states and affect that video or audio signals miss. The obstacle is not the signal but the model. Existing ECG representation models are built for adult clinical data and model assumptions do not hold for single-channel recordings collected over long periods in naturalistic settings.

Recent ECG foundation models are predominantly trained on adult clinical datasets, usually standardized 12-lead recordings of ten seconds from patients in controlled resting conditions~\citep{na2024stmem,mckeen2025ecgfm,nguyen2025tolerantecg,liu2024merl,tian2024ked}.
\begin{wrapfigure}{r}{0.48\linewidth}
    \centering
    \includegraphics[width=\linewidth]{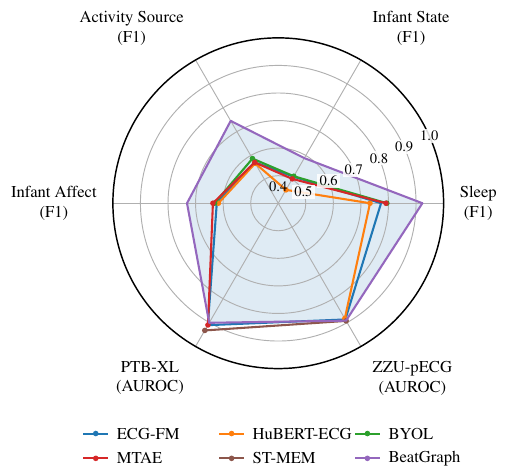}
    \caption{Performance of BeatGraph compared with self-supervised baselines on four infant tasks (macro-F1) and two external benchmarks (AUROC).}
    \label{fig:radar_plot}
\end{wrapfigure}
Some architectures depend on this format directly, modeling the spatial relationships among leads or adapting to 
different lead subsets~\citep{na2024stmem}. Infant ECG falls outside this distribution; autonomic regulation is still maturing through the first years of life \citep{harteveld2021}, so resting heart rate is close to twice the adult rate \citep{fleming2011,rijnbeek2001}. Beat morphology differs as well: the infant heart is right-ventricular dominant and shifts toward the adult pattern over the first months, and the PR, QRS, and QT intervals are all substantially shorter, so waveform features that adult criteria treat as abnormal are normal at this age \citep{rijnbeek2001,davignon1979}. Recording conditions also differ, and it is here that the mismatch with clinical ECG is widest. Wearable home sensors use fewer leads by design, to keep the device small and comfortable for an infant~\citep{islam2024littlebeats}, which removes the 12-lead structure that ECG foundation models typically rely on. Fine-tuning adult models does not address these differences, because the mismatch is in what the model was built to represent.

Pretraining on infant ECG data provides a clear solution given that human labels of infant behavioral and affect states (e.g., sleeping, crying) are costly to collect while infant wearables can produce hundreds of thousands of unlabeled segments during ordinary activity. The open question is how to present that signal to the model. Most self-supervised time-series and ECG models divide a recording into fixed-length patches and reconstruct or contrast them ~\citep{nie2023patchtst,zerveas2021tst,na2024stmem,kiyasseh2021clocs,mehari2022ssl}. Fixed-length patches do not reflect cardiac structure. A patch boundary can fall in the middle of a heartbeat, and the number of beats inside a patch changes with heart rate, which for the fast infant rate means a patch-based model must recover the beat structure before it can reason about relationships between beats.

We introduce \textbf{BeatGraph}, which makes the heartbeat the unit of representation. For each thirty-second window, R-peak detection is used to locate individual beats, and each beat becomes a node in a graph. This approach directly answers the tokenization problem because the model no longer has to recover beat structure from arbitrary patches. A shared beat encoder turns each beat into a feature vector from its waveform and its inter-beat intervals. Thus, both morphology and rate variability are captured at the node level, which are the two channels through which cardiac information is expressed. A Transformer with positional encoding then orders the beats in time, and residual graph attention layers~\citep{vaswani2017attention,velickovic2018gat, brody2022gatv2} operate over all beat pairs, so the model can connect similar cardiac cycles that are not adjacent in the window. An attention-based pooling step then combines the beats into a single window-level representation. We pretrain BeatGraph on the unlabeled infant corpus by masking a subset of beats and predicting their embeddings from the surrounding beats and a self-distillation regularizer~\citep{grill2020byol}, then fine-tune it with task-specific heads for downstream tasks.

To train and evaluate BeatGraph, we collected a corpus of single-channel ECG from an infant wearable sensor used in home, classrooms, and laboratory settings. Our experiments show that this single infant-pretrained encoder transfers across tasks and across populations. BeatGraph is fine-tuned for four infant tasks: sleep-wake state classification, four-class infant state classification, infant- or caregiver-initiated movement, and infant affect recognition, reaching an F1-score of 0.589 to 0.923.  We further evaluate whether the learned representation transfers beyond the population and recording environment used for pretraining. On the ZZU-pECG pediatric benchmark, BeatGraph achieves comparable performance, despite pretraining exclusively on home recordings from infants aged 3 to 11 months. On PTB-XL, it matches ST-MEM under linear evaluation, though it trails after full fine-tuning. Pretrained on adult ECG instead, the same architecture leads all compared methods under linear evaluation and reaches 0.825 against 0.804 for ST-MEM in the single-channel setting.  

Development of infant ECG models has been held back by the data itself. Open infant ECG recordings are scarce; the ones that exist come from small clinical samples (e.g., hospital monitoring of ten preterm infants~\citep{gee2017pics}), and larger pediatric datasets are drawn from sleep laboratories~\citep{lee2022nch}. Neither setting captures continuous recordings of infants in their natural environments, such as home, and neither carries infant states and affect labels that are central to assessing early developmental processes. We release the corpus used in this work to close that gap. It contains 3,408 hours of ECG from 143 infants aged 3 to 11 months, of which 3,361 hours are unlabeled for pretraining and 47.5 hours carry infant state and affect labels for the benchmark tasks. To our knowledge, it is the first public corpus of infant ECG collected in naturalistic settings with behavioral state and affect labels, and supports pediatric representation learning without adapting adult pipelines.

\begin{figure*}[!htb]
    \centering
    \includegraphics[width=0.95\linewidth]{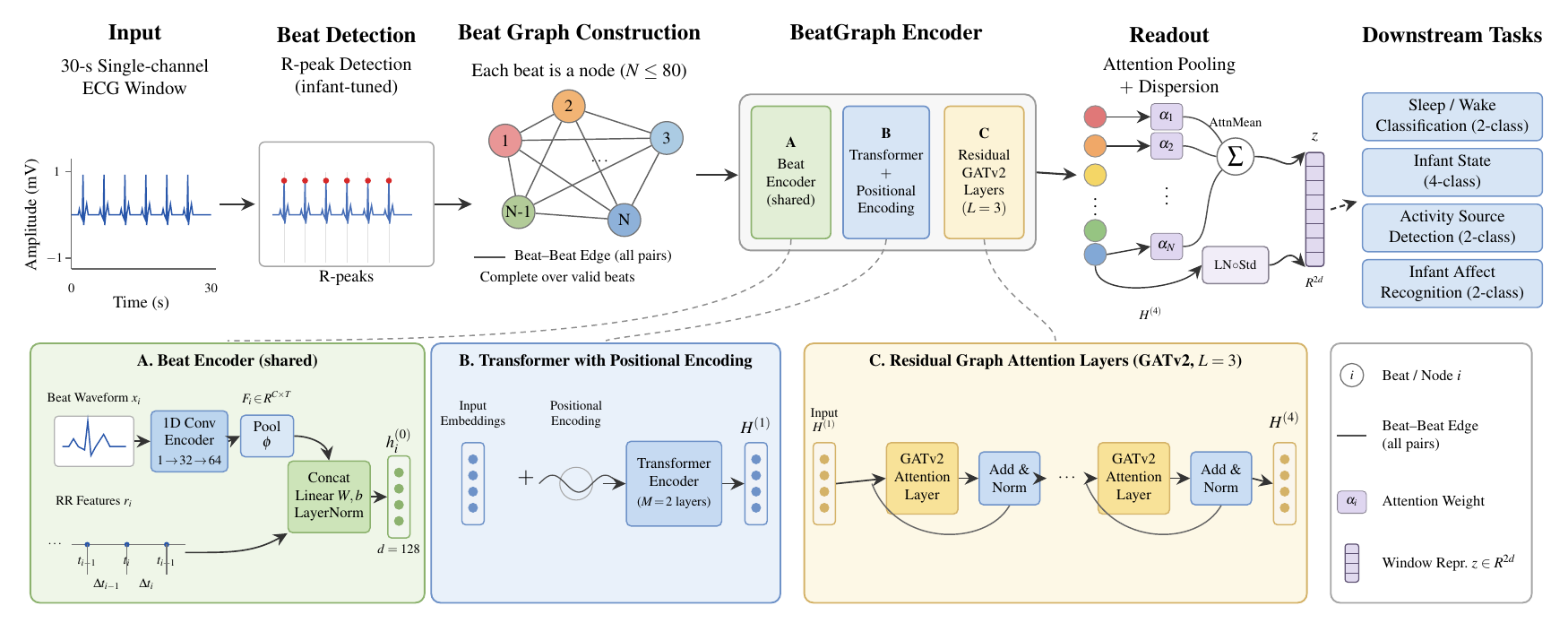}
    \caption{Overview of BeatGraph. R-peak detection converts each ECG window into a complete graph of heartbeats, encoded by a shared beat encoder, a temporal Transformer, and residual GATv2 layers, then pooled by variability-preserving attention into a window embedding. The model is pretrained with masked beat-embedding prediction and self-distillation.}
    \label{fig:main_architecture}
\end{figure*}

\noindent\textbf{Contributions.}
\begin{itemize}
\item We introduce BeatGraph, a self-supervised encoder that represents an ECG window as a graph of individual heartbeats rather than fixed-length patches. The architecture combines beat-level waveform and interval encoding, sequential context from a Transformer, graph attention over beats, and pretraining by masked beat-embedding prediction with a self-distillation regularizer.

\item We show that a single infant-pretrained encoder transfers across tasks and populations, supporting four infant downstream tasks, matching the best published ECG model on the pediatric ZZU-pECG benchmark, and matching adult foundation models on PTB-XL under linear evaluation without adult ECG in pretraining.

\item We evaluate BeatGraph exhaustively across infant, pediatric, and adult ECG datasets and provide ablations examining the contributions of beat detection, graph attention, tokenization effect and beat-window capacity.

\item We release the first public corpus of infant ECG recorded in home, daycare classrooms, and laboratory settings and with infant-state and affect annotations, including unlabeled data, benchmark tasks, and subject-level splits.

\end{itemize}

\section{Related Work}
\label{related_work}

\noindent\textbf{ECG foundation models.}
Recent ECG foundation models learn transferable representations from large unlabeled or
weakly labeled corpora: ST-MEM uses spatio-temporal masked reconstruction over lead-by-time
signals \citep{na2024stmem}, ECG-FM combines contrastive and generative objectives
\citep{mckeen2025ecgfm}, HuBERT-ECG adapts masked unit prediction from speech
\citep{hsu2021hubert,coppola2024hubertecg}, and TolerantECG targets noise and missing leads
\citep{nguyen2025tolerantecg}. Multimodal variants align ECG with reports or language-derived
knowledge \citep{liu2024merl,tian2024ked}. All target adult, clinical, multi-lead ECG and
tokenize raw waveform patches or leads.

\noindent\textbf{Self-supervised and graph learning for physiological signals.}
Contrastive, self-distillation, and masked-modeling methods
\citep{chen2020simclr,grill2020byol,he2022mae,zerveas2021tst,nie2023patchtst,zhang2022maefe}
have been applied to physiological signals; ECG methods such as CLOCS and CMSC \citep{kiyasseh2021clocs}
and prior SSL benchmarks \citep{mehari2022ssl} operate on samples, segments, or patches. Prior
graph-based ECG methods build graphs from signal samples, engineered-feature correlations, or
topological transforms \citep{oliveira2024visibility,han2024correlation,zeinalipour2022gnnecg},
typically for supervised arrhythmia classification.

\noindent\textbf{Pediatric and infant data.}
Public infant ECG is scarce, covering preterm NICU monitoring \citep{gee2017pics} and pediatric polysomnography \citep{lee2022nch}. ZZU-pECG collects 14{,}190 recordings from hospitalized children \citep{tan2025zzupecg}. All are clinical, and neither carries developmentally relevant state labels. Our data come from a multimodal infant wearable \citep{islam2024littlebeats} that can be worn at home, daycare, or laboratory settings, supporting evaluation across sleep, autonomic, behavioral, and affective states.

\noindent\textbf{Our position.}
BeatGraph makes the heartbeat the token: nodes are individual beats, the edge set is unrestricted so attention learns which beat-to-beat relations matter rather than fixing them in advance, and pretraining is masked beat-embedding prediction with BYOL-style self-distillation
on unlabeled single-channel infant recordings, without adult pretraining, clinical reports, or labels.

\section{Method}

\label{sec:method}
BeatGraph encodes a 30-second window as a graph with one node per heartbeat, mapping it to a single window-level embedding in three stages. A beat encoder produces the node features, a temporal Transformer with positional encoding orders the beats, and a graph-attention stage relates them and pools the result. Figure~\ref{fig:main_architecture} depicts the model architecture.

\subsection{Beat-graph construction}
\label{sec:graph-construction}

R-peaks are located with NeuroKit~\citep{makowski2021neurokit} under an infant-tuned
configuration. Beat $i$ is the slice centred on its R-peak, spanning $\mathrm{pre}$ samples before and $\mathrm{post}$ after, both derived from the window's median RR interval $\widetilde{\mathrm{RR}}$ (in ms) and
clipped to fixed bounds,
\begin{equation}
  pre = \big[0.38\,\widetilde{RR}\big]_{130}^{400}
  \qquad
  post = \big[0.52\,\widetilde{RR}\big]_{220}^{600}
  \label{eq:crop}
\end{equation}
At $1000$\,Hz ($1$\,ms${\,=\,}1$ sample) each beat spans $L=\mathrm{pre}+\mathrm{post}$
samples; realised crops are $350$--$675$ samples (median $384$) at a mean $\widetilde{\mathrm{RR}}$ of $438$\,ms ($137$\,bpm). All beats in a window share $(\mathrm{pre},\mathrm{post})$, so the beat tensor is rectangular for batched convolution while $L$ varies across windows. Adjacent crops overlap for $5.1\%$ of beat pairs (above $171$\,bpm), and all but $0.02\%$ stop short of the neighbouring R-peak, so each node holds one centred QRS complex. Node $v_i$ carries the waveform $x_i\in\mathbb{R}^{L}$ and a timing vector
$r_i=[\mathrm{RR}^{\text{prev}}_i,\ \mathrm{RR}^{\text{next}}_i]\in\mathbb{R}^{2}$ in seconds.
We keep up to $N=80$ beats per window, a cap at $160$\,bpm that $1.7\%$ of windows exceed;
these are truncated to the first $N$ beats. Windows are zero-padded to $N$ with a validity mask
$m\in\{0,1\}^{N}$ that excludes padded positions from attention, pooling and all losses. The graph is complete over valid beats, $\mathcal{N}(i)=\{j:m_j=1\}$.

\subsection{Encoder Architecture}

\noindent\textbf{Beat Encoder.} A shared 1-D convolutional stack ($1\!\to\!32\!\to\!64$ channels with downsampling) maps the beat waveform to
a feature map $F_i=\mathrm{Conv}(x_i)\in\mathbb{R}^{C\times T}$ with $C=64$ and $T\approx L/4$,
whose temporal extent inherits the window-dependent crop length. A pooling operator
$\phi:\mathbb{R}^{C\times T}\!\rightarrow\!\mathbb{R}^{d_\phi}$ then removes the time axis; the
default model uses global average pooling, $\phi(F_i)=\tfrac{1}{T}\sum_{t=1}^{T}F_i[:,t]\in\mathbb{R}^{64}$. The pooled descriptor is concatenated with the timing features, projected to the model width $d=128$, and layer
normalised, giving the initial node feature
\begin{equation}
  h^{(0)}_i = \mathrm{LN}\big(W[\,\phi(F_i)\,\|\,r_i\,]+b\big)\in R^{d},
\end{equation}

\noindent\textbf{Temporal Transformer with positional encoding}
Sinusoidal positional encodings $\mathrm{PE}$ are added to the beat sequence, and a shallow
pre-norm Transformer encoder with $2$ layers and $4$ heads~\citep{vaswani2017attention}
contextualises the beats in time, with padded positions excluded through $m$:
\begin{equation}
  H^{(1)} = \mathrm{LN}\!\left(\mathrm{TransformerEnc}\!\left(H^{(0)}+\mathrm{PE},\; m\right)\right).
  \label{eq:temporal}
\end{equation}

\noindent\textbf{Graph Attention.} Three residual graph-attention layers refine the contextualised beats. Suppressing layer and
head indices and writing $h_i$ for $h^{(\ell)}_i$, with per-head width $d'=d/K=32$,
$W_l,W_r\in\mathbb{R}^{d'\times d}$ and $a\in\mathbb{R}^{d'}$, we use the GATv2
score~\citep{brody2022gatv2},
\begin{equation}
  e_{ij} = a^{\top}\mathrm{LeakyReLU}\big(W_l h_i + W_r h_j\big),
\end{equation}
normalized over neighbourhood as $\alpha_{ij}=\mathrm{softmax}_j\, e_{ij}$ and aggregated as
$u_i=\sum_{j\in\mathcal{N}(i)}\alpha_{ij}W_r h_j$, with negative slope $0.2$ and
padded nodes excluded from $\mathcal{N}(i)$. The $K=4$ heads are concatenated, projected by
$W_o$, and added residually,
\begin{equation}
  h^{(\ell+1)}_i = \mathrm{LN}\big(h^{(\ell)}_i + W_o[\,u^{(1)}_i\|\cdots\|u^{(K)}_i\,]\big),
\end{equation}

\noindent\textbf{Readout} Valid beats are pooled into the window embedding by an attention-weighted mean.  A plain mean discards how much beats differ from one another, which is where heart-rate variability resides, so we concatenate a scale-normalised second-order term,
\begin{equation}
  z = \big[\,\mathrm{AttnMean}(H,m)\,\big\|\,\mathrm{LN}\,\mathrm{Std}(H,m)\,\big]\in R^{2d},
\end{equation}
writing $H$ for $H^{(4)}$, with both statistics over valid beats only.

\subsection{Self-supervised Pretraining}
Pretraining combines masked beat-embedding prediction with a self-distillation regularizer and uses no
labels. Targets come from a separate pass over the unmasked sequence with gradients blocked,
$h^{(\mathrm{tgt})}=\mathrm{sg}\big[H^{(1)}\big]$. A random fraction $\rho=0.3$ of the valid
beats, $\mathcal{M}\subset\{i:m_i=1\}$, is replaced by a learned mask token before the temporal
Transformer, with positional encodings added after masking so each token retains the identity
of the beat it must predict; the masked sequence then passes through the Transformer, the graph
layers and a linear decoder $g$, giving $\hat h_i=g\big(h^{(4)}_i\big)$. Following
BYOL~\citep{grill2020byol}, $\mathcal{L}_{\mathrm{BYOL}}$ is a symmetric cosine loss between an
online network and an exponential-moving-average target (decay $0.99$) on two augmented views,
which discourages the constant-embedding solutions that a same-weights latent target admits.
The objective is
\begin{equation}
  \mathcal{L} = \frac{1}{|\mathcal{M}|}\sum_{i\in\mathcal{M}}
     \big\|\hat h_i - h^{(\mathrm{tgt})}_i\big\|_2^2
     + \lambda_{\mathrm{BYOL}}\,\mathcal{L}_{\mathrm{BYOL}},
\end{equation}
with $\lambda_{BYOL}=0.5$. The first term is the masked prediction loss. We train for a fixed budget of 100 epochs and use the final checkpoint.

\subsection{Downstream Fine-Tuning}
For each labeled task, we fine-tune the pretrained encoder together with a task head, a residual MLP, on 30-second windows, and we train for a fixed 20 epochs and use the final checkpoint. For multi-lead external datasets the shared single-channel beat encoder is applied to each lead, and the per-lead embeddings are concatenated across the 12 leads before the classification head, with any missing leads zero-filled. External records are shorter than the $30$\,s infant windows but need no re-windowing as each record forms one graph and the validity mask absorbs the smaller beat count, so $N=80$ is a capacity bound rather than a fixed length, and Eq.~\eqref{eq:crop} applies unchanged. The same infant-trained encoder is reused without any architectural change except the head's input width. For the evaluation on external data, we provide linear evaluation with a $nn.Linear$ head keeping the encoder frozen, and fine-tuned evaluation, where we fine-tune the encoder too to match the baseline evaluation.

\subsection{Implementation details}
Recordings are resampled to $1000$\,Hz, segmented into non-overlapping $30$\,s windows, and
z-scored per window. The model uses $d=128$, $K=4$ heads of width $d'=32$, two Transformer
layers and three graph layers, totalling 1.37 \,M parameters in the encoder. The BYOL target network and projection heads are discarded after pretraining. Pretraining runs for $100$ epochs over the unlabelled windows using AdamW at learning rate $10^{-3}$, weight decay $10^{-4}$, $5$ warm-up epochs then
cosine decay to a $10^{-5}$ floor, and EMA decay $0.99$.

\section{The Infant ECG Corpus}
\label{sec:data}

\subsection{Sensing platform and cohorts}
We use \emph{$LittleBeats^{TM}$}, a lightweight infant wearable platform that integrates a 3-lead single-channel ECG, an inertial measurement unit, and a microphone. The device is secured in the front pocket of a specially designed onesie worn by the infant, and parents report that the device is easy to use and does not interfere with infant activities during daylong recordings \citep{mcelwain2024evaluating}. Each recording is a daylong session, meaning 8-10 hours of continuous single-channel ECG sampled at 1000 Hz. The corpus comprises two cohorts. A large \emph{unlabeled} cohort (used for self-supervised pretraining) consists of 3361 hours of recordings from 139 infants. A smaller \emph{labeled} cohort of 39.3 hours from four infants supports the first three downstream tasks described below. Infants in this second, smaller cohort were recruited from child care centers in the community and wore LittleBeats in their infant classrooms 1-2 times per week over the course of 7-15 weeks. At the beginning of each classroom recording, a trained Research Assistant (RA) conducted live observations of relevant infant behaviors (described below) while simultaneously video recording the infant with a chest-mounted GoPro camera. To ensure high accuracy of the live observations, a second trained RA subsequently reviewed and updated the annotations as needed, with the video recording as the reference point. For the fourth task below, we used data from 122 infants from the first cohort who participated in a mother-infant visit to our laboratory playroom.

\subsection{Pretraining data}
For pretraining, we use approximately 3361 hours of ECG data recorded in the home. Each recording is segmented into non-overlapping 30-second windows, and heartbeats are located by the infant-tuned NeuroKit configuration of the previous section. No labels are used at this stage, and no labeled recording enters the pretraining corpus. The labeled classroom recordings come from a cohort disjoint from the pretraining cohort, and the affect sessions are separate laboratory recordings.

\subsection{Downstream tasks}
We define four labeled tasks that span developmentally relevant states an infant wearable can plausibly monitor. We use \emph{subject-aware} split, maintaining leave-one-infant-out protocol on first three tasks. For the infant affect recognition task, the held-out infants contribute unlabeled data to the pretraining corpus, so that the result measures generalization to unseen labels rather than unseen infants.

\noindent\textbf{Sleep/wake (2-class).}
Two-way classification of sleep/wake, representing the most basic axis of infant state regulation.

\noindent\textbf{Infant states (4-class).}
A four-way classification of infant states: (a) crying/fussing as indicated by vocal and nonvocal cues of distress or discomfort, (b) quiet alert as indicated by open eyes, relaxed body with minimal movement, and steady breathing, (c) active alert as indicated by a moderate to high degree of body movement, vocalizations, and/or exploration of objects, and (d) sleeping. These four states are likely to exhibit salient differences in heart-rate-variability information and thus serve as our canonical \emph{variability} task.

\noindent\textbf{Activity source (2-class)}
Two-way classification of whether infant movement originates with the infant or is influenced by the caregiver. Infant-initiated movement included periods when the infant was on the floor, crying or actively alert, and not in physical contact with a caregiver; it could also include  independent movements by the infant, such as crawling or scooting. Caregiver-initiated movement included periods when a caregiver touched the infant’s torso, manipulated the infant’s limbs, or moved the infant through space by carrying, swaying, or rocking. Self-generated activity engages the infant's own motor and autonomic systems, whereas caregiver-generated movement displaces the infant without that central drive, so the two conditions differ in cardiac response. We have 22.2 hours of labeled activity source from four infants.

\noindent\textbf{Infant Affect Recognition (2-class).}
For the final task, we use data from the Still-Face Paradigm \citep{tronick1978stillface}, which includes three 2-minute episodes: face-to-face play, maternal still face, and reunion. We use the play and still-face episode labels as proxies for infants’ positive and negative affect, respectively. Maternal disengagement during the still-face episode disrupts expected reciprocal interaction and reduces positive affect while increasing negative affect \citep{mesman2009stillface}.We used 8.2 hours of data from 122 infant-mother dyads.

\subsection{Released benchmark}
We release the unlabeled and labeled raw and processed ECG together with task definitions, subject-level train/validation/test splits, and code. The pretraining data and the sleep/wake, infant states, and activity source annotations come from home and classroom recordings, while infant affect labels come from structured lab visits. To our knowledge, this is the first public infant ECG resource that is (i) recorded in naturalistic settings rather than clinics, (ii) continuous rather than short strips, and (iii) annotated across multiple behavioral states, covering sleep/wake, infant states, activity source, and affect, for children aged 3 to 11 months.

\begin{table}[b]
\centering
\caption{Results on the four infant downstream tasks. All methods are
pretrained on the same infant corpus. BeatGraph achieves the strongest
performance across every task and evaluation metric. Bold denotes the best
result, and underlining denotes the second-best result.}
\label{tab:infant_downstream_results}
\setlength{\tabcolsep}{3.2pt}
\adjustbox{max width=\linewidth}{%
\begin{tabular}{llcccccc}
\toprule
Task & Metric
& ECG-FM
& HuBERT-ECG
& BYOL
& SimCLR
& MTAE
& BeatGraph \\
\midrule

\multirow{3}{*}{\shortstack{Sleep\\Classification}}
& Acc.
& $0.832 \pm 0.023$
& $0.793 \pm 0.025$
& $0.852 \pm 0.025$
& $0.781 \pm 0.028$
& $\underline{0.858 \pm 0.031}$
& $\mathbf{0.941 \pm 0.026}$ \\

& F1
& $0.773 \pm 0.035$
& $0.733 \pm 0.043$
& $\underline{0.794 \pm 0.048}$
& $0.737 \pm 0.041$
& $0.791 \pm 0.052$
& $\mathbf{0.923 \pm 0.038}$ \\

& $\kappa$
& $0.583 \pm 0.062$
& $0.530 \pm 0.057$
& $\underline{0.611 \pm 0.068}$
& $0.514 \pm 0.061$
& $0.607 \pm 0.068$
& $\mathbf{0.812 \pm 0.057}$ \\
\midrule

\multirow{3}{*}{\shortstack{Infant State\\Detection}}
& Acc.
& $\underline{0.677 \pm 0.023}$
& $0.647 \pm 0.015$
& $0.652 \pm 0.022$
& $0.623 \pm 0.027$
& $0.648 \pm 0.015$
& $\mathbf{0.732 \pm 0.023}$ \\

& F1
& $0.511 \pm 0.035$
& $0.456 \pm 0.018$
& $\underline{0.513 \pm 0.033}$
& $0.512 \pm 0.023$
& $0.502 \pm 0.028$
& $\mathbf{0.589 \pm 0.027}$ \\

& $\kappa$
& $\underline{0.431 \pm 0.044}$
& $0.359 \pm 0.032$
& $0.423 \pm 0.049$
& $0.404 \pm 0.057$
& $0.425 \pm 0.042$
& $\mathbf{0.483 \pm 0.038}$ \\
\midrule

\multirow{3}{*}{\shortstack{Activity Source\\Recognition}}
& Acc.
& $0.631 \pm 0.017$
& $0.615 \pm 0.015$
& $0.633 \pm 0.019$
& $0.612 \pm 0.018$
& $\underline{0.634 \pm 0.013}$
& $\mathbf{0.767 \pm 0.015}$ \\

& F1
& $0.572 \pm 0.018$
& $0.567 \pm 0.029$
& $\underline{0.587 \pm 0.022}$
& $0.563 \pm 0.021$
& $0.571 \pm 0.015$
& $\mathbf{0.745 \pm 0.021}$ \\

& $\kappa$
& $0.313 \pm 0.034$
& $0.272 \pm 0.031$
& $\underline{0.338 \pm 0.051}$
& $0.268 \pm 0.052$
& $0.317 \pm 0.029$
& $\mathbf{0.513 \pm 0.031}$ \\
\midrule

\multirow{3}{*}{\shortstack{Infant Affect\\Recognition}}
& Acc.
& $0.721 \pm 0.013$
& $0.714 \pm 0.012$
& $0.733 \pm 0.015$
& $0.731 \pm 0.015$
& $\underline{0.742 \pm 0.013}$
& $\mathbf{0.812 \pm 0.014}$ \\

& F1
& $0.623 \pm 0.023$
& $0.617 \pm 0.022$
& $0.631 \pm 0.025$
& $\underline{0.639 \pm 0.015}$
& $0.637 \pm 0.017$
& $\mathbf{0.731 \pm 0.019}$ \\

& $\kappa$
& $0.252 \pm 0.019$
& $0.223 \pm 0.028$
& $0.273 \pm 0.033$
& $0.254 \pm 0.032$
& $\underline{0.278 \pm 0.036}$
& $\mathbf{0.423 \pm 0.023}$ \\
\bottomrule
\end{tabular}}

\end{table}
\begin{table}[t]
\centering
\caption{PTB-XL results under linear evaluation and full fine-tuning. BeatGraph-Infant and BeatGraph-Adult denote pretraining on infant and adult ECG, respectively. Bold denotes the best result, and underlining denotes the second-best result.}
\label{tab:ptbxl_linear_finetune}
\setlength{\tabcolsep}{3pt}
\adjustbox{max width=\linewidth}{%
\begin{tabular}{lccc ccc}
\toprule
& \multicolumn{3}{c}{Linear Evaluation}
& \multicolumn{3}{c}{Full Fine-Tuning} \\
\cmidrule(lr){2-4}
\cmidrule(lr){5-7}
Method
& Accuracy & F1 & AUROC
& Accuracy & F1 & AUROC \\
\midrule
MoCo v3
& $0.552 \pm 0.000$
& $0.142 \pm 0.000$
& $0.739 \pm 0.006$
& $0.799 \pm 0.004$
& $0.644 \pm 0.010$
& $0.913 \pm 0.002$ \\

CMSC
& $0.681 \pm 0.032$
& $0.441 \pm 0.051$
& $0.797 \pm 0.038$
& $0.724 \pm 0.067$
& $0.510 \pm 0.111$
& $0.877 \pm 0.003$ \\

MTAE
& $0.683 \pm 0.008$
& $0.437 \pm 0.012$
& $0.807 \pm 0.006$
& $0.789 \pm 0.002$
& $0.613 \pm 0.012$
& $0.910 \pm 0.001$ \\

ECG-FM
& $0.692 \pm 0.005$
& $0.451 \pm 0.011$
& $0.796 \pm 0.007$
& $0.788 \pm 0.004$
& $0.609 \pm 0.009$
& $0.911 \pm 0.004$ \\

MLAE
& $0.649 \pm 0.008$
& $0.382 \pm 0.020$
& $0.779 \pm 0.008$
& $0.802 \pm 0.004$
& $0.625 \pm 0.009$
& $0.915 \pm 0.001$ \\

TolerantECG
& $0.671 \pm 0.006$
& $0.456 \pm 0.008$
& $0.801 \pm 0.011$
& $0.812 \pm 0.006$
& $0.635 \pm 0.010$
& $0.922 \pm 0.006$ \\

ST-MEM
& $0.726 \pm 0.005$
& $0.508 \pm 0.008$
& $0.838 \pm 0.013$
& $\mathbf{0.825 \pm 0.002}$
& $\mathbf{0.655 \pm 0.003}$
& $\mathbf{0.933 \pm 0.003}$ \\
\midrule

BeatGraph-Infant
& $\underline{0.731 \pm 0.004}$
& $\underline{0.513 \pm 0.005}$
& $\underline{0.842 \pm 0.008}$
& $0.806 \pm 0.003$
& $0.633 \pm 0.004$
& $0.901 \pm 0.006$ \\

BeatGraph-Adult
& $\mathbf{0.780 \pm 0.003}$
& $\mathbf{0.540 \pm 0.005}$
& $\mathbf{0.870 \pm 0.007}$
& $\underline{0.823 \pm 0.002}$
& $\underline{0.651 \pm 0.004}$
& $\underline{0.931 \pm 0.004}$ \\
\bottomrule
\end{tabular}}

\end{table}

\subsection{External datasets for transfer evaluation}
We evaluate on public pediatric and adult ECG under their standard patient-level protocols, and pretrain the same recipe on adult data to test the reverse direction. \textbf{ZZU-pECG} \citep{tan2025zzupecg} contributes 14{,}190 pediatric records from 11{,}643 children aged 0 to 14 years, and \textbf{PTB-XL} \citep{wagner2020ptbxl} contributes 21{,}799 adult 12-lead records from 18{,}869 patients under the standard Wagner splits and diagnostic hierarchy. \textbf{MIMIC-IV-ECG} \citep{gow2023mimicivecg} serves as the adult pretraining source, roughly 2{,}200 hours across 800k records and comparable in scale to our infant corpus, with PTB-XL held out for evaluation.

\section{Evaluation}
\label{sec:experiments}

\paragraph{Baselines and Evaluation Setup.} 
We compare BeatGraph against representative waveform-based self-supervised approaches, ECG-FM, HuBERT-ECG, SimCLR, BYOL, and MTAE \citep{zhang2022maefe}. All baselines follow their published implementations apart from taking single-channel input, and each is pretrained on the same unlabeled infant corpus as BeatGraph, with tuning details in Appendix~\ref{app:implementation}. On the three classroom tasks, we use leave-one-infant-out cross-validation, with each infant serving as the test set in turn and each fold run with five seeds. We average seeds within a fold and report the mean and standard deviation across folds, so the reported spread reflects variation between infants rather than between seeds. The affect recognition task uses a subject-level split across the 122 dyads with five seeds. As noted in the corpus description, these infants contribute unlabeled data to pretraining, so the affect recognition result measures generalization to unseen labels rather than unseen subjects. Fine-tuning hyperparameters were fixed a priori and were not tuned on infant data. For external transfer, we compare against MoCo v3 \citep{chen2021empirical}, CMSC, MTAE, ECG-FM, MLAE \citep{zhang2022maefe}, Tolerant-ECG, and ST-MEM on PTB-XL under their linear-evaluation and fine-tuning protocols. PTB-XL results for most baselines are taken from the ST-MEM paper as published, and we regenerate ECG-FM and Tolerant-ECG from their released checkpoints. ZZU-pECG baseline results are reported as published. Supervised deep learning baselines and tree-based classifiers trained on heart-rate-variability features are in Appendix~\ref{app:physio-baselines}.

\begin{table}[t]
\centering
\begin{minipage}[t]{0.47\linewidth}

\centering\small
\setlength{\tabcolsep}{7pt}
\caption{PTB-XL AUROC for BeatGraph-Adult under twelve-lead and single-lead evaluation. Both models are pretrained on adult ECG and fine-tuned on PTB-XL.}
\label{tab:ptbxl_lead_comparison}
\adjustbox{max width=\linewidth}{%
\begin{tabular}{lcc}
\toprule
Method & 12-Channels & 1-Channel \\
\midrule
ST-MEM
& $\mathbf{0.933 \pm 0.003}$
& $0.804 \pm 0.005$ \\
BeatGraph-Adult
& $0.931 \pm 0.004$ 
& $\mathbf{0.825 \pm 0.005}$ \\
\bottomrule
\end{tabular}}
\end{minipage}\hfill
\begin{minipage}[t]{0.49\linewidth}
\setlength{\tabcolsep}{1mm}
\caption{AUROC on the ZZU-pECG pediatric benchmark across 58 diagnostic
outputs \citep{tan2025zzupecg}. Bold denotes the best, and
underlining denotes the second-best result.}
\label{tab:zzu_auroc}
\adjustbox{max width=\linewidth}{%
\begin{tabular}{@{}lccccc@{}}
\toprule
Metric
& ECG-FM
& MERL
& \shortstack{HuBERT\\-ECG}
& \shortstack{ST\\-MEM}
& \shortstack{BeatGraph} \\
\midrule

AUROC
& 0.887
& 0.886
& 0.883
& \textbf{0.893}
& \underline{0.892} \\

\bottomrule
\end{tabular}}

\end{minipage}
\end{table}

\begin{table}[t]
\centering
\begin{minipage}[t]{0.49\linewidth}
\centering\small
\setlength{\tabcolsep}{2pt}
\caption{Performance comparison of Adult-pretrained ECG models fine-tuned on the infant downstream
tasks with infant-pretrained BeatGraph.}
\label{tab:adult_to_infant_transfer}
\adjustbox{max width=\linewidth}{%
\begin{tabular}{@{}lcccc@{}}

\toprule
Method & Sleep & State & Source & Affect \\
\midrule

ECG-FM
& $0.70_{\pm 0.06}$
& $0.41_{\pm 0.02}$
& $0.50_{\pm 0.02}$
& $0.51_{\pm 0.04}$ \\

MTAE
& $0.68_{\pm 0.05}$
& $0.39_{\pm 0.01}$
& $0.51_{\pm 0.03}$
& $0.51_{\pm 0.04}$ \\

BYOL
& $0.69_{\pm 0.05}$
& $0.40_{\pm 0.04}$
& $0.49_{\pm 0.03}$
& $0.50_{\pm 0.03}$ \\

ST-MEM
& $0.68_{\pm 0.06}$
& $0.39_{\pm 0.04}$
& $0.52_{\pm 0.02}$
& $0.52_{\pm 0.04}$ \\

\midrule

\textbf{BeatGraph}
& $\mathbf{0.92}_{\pm 0.04}$
& $\mathbf{0.59}_{\pm 0.03}$
& $\mathbf{0.75}_{\pm 0.02}$
& $\mathbf{0.73}_{\pm 0.02}$ \\

\bottomrule
\end{tabular}}
\medskip
\centering\small
\setlength{\tabcolsep}{1mm}
\caption{Effect of Transformer and graph-attention components on downstream
macro-F1. T denotes Transformer, and 32-kNN connects each beat to its 32 nearest neighbors rather than using the fully connected graph of the proposed model.}
\label{tab:ablation_graph}
\adjustbox{max width=\linewidth}{%
\begin{tabular}{@{}lcccc@{}}
\toprule
Configuration & Sleep & State & Source & Affect \\
\midrule

2T w/o GAT
& $0.78_{\pm 0.07}$
& $0.46_{\pm 0.04}$
& $0.58_{\pm 0.04}$
& $0.57_{\pm 0.03}$ \\

6T w/o GAT
& $0.81_{\pm 0.05}$
& $0.52_{\pm 0.03}$
& $0.63_{\pm 0.03}$
& $0.61_{\pm 0.04}$ \\

GAT w/o T
& $0.83_{\pm 0.06}$
& $0.53_{\pm 0.04}$
& $0.64_{\pm 0.03}$
& $0.62_{\pm 0.03}$ \\

32-kNN
& $\underline{0.89}_{\pm 0.05}$
& $\underline{0.57}_{\pm 0.03}$
& $\mathbf{0.76}_{\pm 0.03}$
& $\underline{0.72}_{\pm 0.03}$ \\

\textbf{Proposed}
& $\mathbf{0.92}_{\pm 0.04}$
& $\mathbf{0.59}_{\pm 0.03}$
& $\underline{0.75}_{\pm 0.02}$
& $\mathbf{0.73}_{\pm 0.02}$ \\

\bottomrule
\end{tabular}}
\end{minipage}\hfill
\begin{minipage}[t]{0.49\linewidth}
\centering\small
\setlength{\tabcolsep}{1mm}
\caption{Tokenization effect on downstream macro-F1, with encoder, graph, pretraining, and fine-tuning held fixed. }
\label{tab:tokenization_ablation}
\adjustbox{max width=\linewidth}{%
\begin{tabular}{@{}lcccc@{}}
\toprule
Tokenization
& Sleep
& State
& \shortstack{Source}
& Affect \\
\midrule

Fixed patch
& $0.78_{\pm 0.04}$
& $0.43_{\pm 0.02}$
& $0.56_{\pm 0.03}$
& $0.59_{\pm 0.03}$ \\
Beat wo. RR
& $0.88_{\pm 0.05}$
& $0.53_{\pm 0.03}$
& $0.65_{\pm 0.02}$
& $0.66_{\pm 0.03}$ \\

\textbf{Beat w. RR}
& $\mathbf{0.92}_{\pm 0.04}$
& $\mathbf{0.59}_{\pm 0.03}$
& $\mathbf{0.75}_{\pm 0.02}$
& $\mathbf{0.73}_{\pm 0.02}$ \\

\bottomrule
\end{tabular}}
\medskip
\centering\small
\setlength{\tabcolsep}{1mm}
\caption{Effect of the maximum number of beats per ECG window on downstream macro-F1.}
\label{tab:ablation_max_beats}
\adjustbox{max width=\linewidth}{%
\begin{tabular}{@{}ccccc@{}}
\toprule
Max Beats & Sleep & State & Source & Affect \\
\midrule

40
& $0.76_{\pm 0.06}$
& $0.46_{\pm 0.05}$
& $0.62_{\pm 0.04}$
& $0.58_{\pm 0.03}$ \\

\textbf{80}
& $\mathbf{0.92}_{\pm 0.04}$
& $\mathbf{0.59}_{\pm 0.03}$
& $\mathbf{0.75}_{\pm 0.02}$
& $\mathbf{0.73}_{\pm 0.02}$ \\

120
& $0.91_{\pm 0.04}$
& $0.58_{\pm 0.04}$
& $0.73_{\pm 0.03}$
& $0.73_{\pm 0.03}$ \\

\bottomrule
\end{tabular}}
\smallskip
\centering\small
\setlength{\tabcolsep}{1mm}
\caption{Effect of the R-peak detection strategy on downstream macro-F1.}
\label{tab:ablation_rpeak}
\adjustbox{max width=\linewidth}{%
\begin{tabular}{@{}lcccc@{}}
\toprule
Detector
& Sleep
& State
& \shortstack{Source}
& Affect \\
\midrule

NeuroKit
& $0.87_{\pm 0.04}$
& $0.53_{\pm 0.02}$
& $0.68_{\pm 0.03}$
& $0.67_{\pm 0.02}$ \\

\textbf{Tuned}
& $\mathbf{0.92}_{\pm 0.04}$
& $\mathbf{0.59}_{\pm 0.03}$
& $\mathbf{0.75}_{\pm 0.02}$
& $\mathbf{0.73}_{\pm 0.02}$ \\

\bottomrule
\end{tabular}}
\end{minipage}
\end{table}

\noindent\textbf{Main Results.}
On the four infant tasks (Table~\ref{tab:infant_downstream_results}), BeatGraph outperforms every waveform-based self-supervised baseline on accuracy, macro-F1, and Cohen's $\kappa$. On PTB-XL (Table~\ref{tab:ptbxl_linear_finetune}) the infant-pretrained encoder matches ST-MEM under linear evaluation despite pretraining only on single-channel recordings from infants 3-11 months of age. Both BeatGraph variants are stronger frozen than fine-tuned relative to ST-MEM, so the gains come from representation quality rather than adaptation capacity. Table~\ref{tab:ptbxl_lead_comparison} isolates the cost of ignoring lead geometry. BeatGraph-Adult applies one single-channel encoder to each lead and concatenates the results with no cross-lead interaction, yet reaches $0.931 \pm 0.004$ against$ 0.933 \pm 0.003$, so the spatial structure ST-MEM models is worth 0.002 here. That independence also limits exposure when leads are removed. Going from twelve leads to one costs ST-MEM 0.129 AUROC and BeatGraph-Adult 0.106, and in the single-lead setting BeatGraph-Adult leads by 0.021, more than three times either standard deviation. On ZZU-pECG (Table~\ref{tab:zzu_auroc}) the infant-pretrained encoder reaches 0.892 AUROC, inside the 0.883 to 0.893 band spanned by four models pretrained on large multi-lead clinical corpora. That benchmark is hospital-recorded diagnostic ECG, so the comparison holds across recording settings as well as age.

\noindent\textbf{Ablations.} 
Table~\ref{tab:adult_to_infant_transfer} applies published adult-pretrained models to the infant tasks by placing the single-channel signal in Lead I and zero-filling the rest, and their weak performance shows these checkpoints cannot be used directly on infant wearable ECG, whether from the pretraining domain or the lead adaptation. Table~\ref{tab:infant_downstream_results} controls for both, since every baseline is pretrained on the same infant corpus without adaptation.

Table \ref{tab:ablation_graph} separates sequential from relational modeling. Increasing the Transformer depth from two to six layers improves performance, but both configurations without graph attention remain below the graph-based variants. Moreover, GAT without a Transformer outperforms both no-GAT models, suggesting that explicit relationships among beats provide information beyond additional sequential modeling. The proposed fully connected graph performs best on three of the four tasks, while the 32-knn graph is slightly better for activity-source recognition. Overall, these results support the contribution of graph-based beat interaction, while also indicating that sparse connectivity may be sufficient for some tasks.
Table~\ref{tab:tokenization_ablation} separates the two things beat tokenization provides. Aligning tokens to heartbeats helps by a similar amount on every task, while the RR features added on top help unevenly. Alignment is therefore not merely a way of delivering rhythm information, since it improves tasks that gain little from the intervals themselves. What it supplies is a consistent unit, so the same position in the cardiac cycle falls at the same position in every token regardless of rate.  In Table~\ref{tab:ablation_max_beats} the beat cap acts as a threshold rather than a budget. Forty beats in a 30-second window corresponds to 80 beats per minute, below the resting infant rate, so that setting truncates almost every window rather than trimming a tail. Raising the cap to 120 admits rates that few windows reach and adds mostly padding. 
In Table~\ref{tab:ablation_rpeak} the detector substitution costs a comparable amount on all four tasks, consistent with a segmentation-quality effect that propagates through the representation rather than one specific to any label.

\section{Conclusion}
We introduced BeatGraph, a self-supervised encoder that represents an ECG window as a graph of individual heartbeats rather than fixed-length patches. Pretraining on 3,361 hours of unlabeled infant ECG enables a single encoder to outperform waveform-based baselines on sleep vs wake, infant state, activity source, and infant affect, to remain competitive on the pediatric ZZU-pECG benchmark, and to match adult foundation models on PTB-XL under linear evaluation. Our ablations locate the gain in the tokenization and the connectivity, since beat-aligned tokens improve every task over fixed patches and additional Transformer depth does not recover what removing graph attention costs. Alongside the model we release the corpus behind it with 3,361 hours of unlabeled infant home ECG for pretraining and 47.5 hours annotated for infant state and affect, with subject-level splits and task definitions. Future work will evaluate beat-detection accuracy against manually annotated ECG and extend the framework to multimodal infant sensing.

\subsection*{AI use statement}
We used generative AI tools for brainstorming during the research process, and polishing the writing of the manuscript, including grammar, phrasing, and clarity. All research ideas that emerged from AI-assisted brainstorming were evaluated by the authors against the existing literature before being pursued, and all AI-edited text was reviewed and revised by the authors to ensure it accurately reflects our methods and results. We take responsibility for the final content of this work, including text, claims, and artifacts produced with the aid of generative AI.

\subsection*{Ethics statement}
All data collection was approved by the institutional review board, and a parent or legal guardian gave written informed consent for every infant before recording, including consent for public release of deidentified physiological data. Released data are deidentified, and audio, IMU, and video streams are withheld. Because ECG can act as a biometric, we assess re-identification risk explicitly, and we discuss the limits of the cohort, the risk that affect or activity-source models could be misused to evaluate caregivers, and the fact that BeatGraph is not a clinical device (Appendix~\ref{app:ethics}).

\subsection*{Reproducibility statement}
The beat-graph construction, encoder, pretraining objective, and fine-tuning protocol are specified in Section~\ref{sec:method}, and the full optimisation settings, heads, hardware, and signal-quality filtering are in Appendix~\ref{app:impl}. Baseline implementations, single-channel adaptations, and tuning are described in Appendices~\ref{app:baselines} and~\ref{app:implementation}, with the provenance of every external number in Table~\ref{tab:app-provenance}. External dataset handling is in Appendix~\ref{app:datasets}, and the leave-one-infant-out folds and class supports are in Appendix~\ref{app:lofo}. We release the corpus with task definitions, subject-level splits, and code.

\bibliography{references}

@inproceedings{na2024stmem,
  title     = {Guiding Masked Representation Learning to Capture Spatio-Temporal Relationship of Electrocardiogram},
  author    = {Na, Yeongyeon and Park, Minje and Tae, Yunwon and Joo, Sunghoon},
  booktitle = {International Conference on Learning Representations (ICLR)},
  year      = {2024},
  note      = {arXiv:2402.09450}
}

@inproceedings{liu2024merl,
  title     = {Zero-Shot {ECG} Classification with Multimodal Learning and Test-time Clinical Knowledge Enhancement},
  author    = {Liu, Che and Wan, Zhongwei and Ouyang, Cheng and Shah, Anand and Bai, Wenjia and Arcucci, Rossella},
  booktitle = {International Conference on Machine Learning (ICML)},
  year      = {2024},
  note      = {arXiv:2403.06659}
}

@article{tian2024ked,
  title   = {Foundation model of {ECG} diagnosis: Diagnostics and explanations of any form and rhythm on {ECG}},
  author  = {Tian, Yuanyuan and Li, Zhiyuan and Jin, Yanrui and Wang, Mengxiao and Wei, Xiaoyang and Zhao, Liqun and Liu, Yunqing and Liu, Jinlei and Liu, Chengliang},
  journal = {Cell Reports Medicine},
  volume  = {5},
  number  = {12},
  pages   = {101875},
  year    = {2024},
  doi     = {10.1016/j.xcrm.2024.101875}
}

@article{mckeen2025ecgfm,
  title   = {{ECG-FM}: an open electrocardiogram foundation model},
  author  = {McKeen, Kaden and Masood, Sameer and Toma, Augustin and
             Rubin, Barry and Wang, Bo},
  journal = {JAMIA Open},
  volume  = {8}, number = {5}, pages = {ooaf122}, year = {2025}
}

@inproceedings{nguyen2025tolerantecg,
  title     = {{TolerantECG}: A Foundation Model for Imperfect Electrocardiogram},
  author    = {Nguyen, Huynh Dang and Pham, Trong-Thang and Le, Ngan and Nguyen, Van},
  booktitle = {Proceedings of the ACM International Conference on Multimedia (ACM MM)},
  year      = {2025},
  note      = {arXiv:2507.09887}
}

@inproceedings{chen2020simclr,
  title     = {A Simple Framework for Contrastive Learning of Visual Representations},
  author    = {Chen, Ting and Kornblith, Simon and Norouzi, Mohammad and Hinton, Geoffrey},
  booktitle = {International Conference on Machine Learning (ICML)},
  year      = {2020}
}

@inproceedings{grill2020byol,
  title     = {Bootstrap Your Own Latent: A New Approach to Self-Supervised Learning},
  author    = {Grill, Jean-Bastien and Strub, Florian and Altch{\'e}, Florent and Tallec, Corentin and Richemond, Pierre H. and Buchatskaya, Elena and Doersch, Carl and Pires, Bernardo Avila and Guo, Zhaohan Daniel and Azar, Mohammad Gheshlaghi and Piot, Bilal and Kavukcuoglu, Koray and Munos, R{\'e}mi and Valko, Michal},
  booktitle = {Advances in Neural Information Processing Systems (NeurIPS)},
  year      = {2020}
}

@inproceedings{kiyasseh2021clocs,
  title     = {{CLOCS}: Contrastive Learning of Cardiac Signals Across Space, Time, and Patients},
  author    = {Kiyasseh, Dani and Zhu, Tingting and Clifton, David A.},
  booktitle = {International Conference on Machine Learning (ICML)},
  year      = {2021}
}

@article{mehari2022ssl,
  title   = {Self-supervised representation learning from 12-lead {ECG} data},
  author  = {Mehari, Temesgen and Strodthoff, Nils},
  journal = {Computers in Biology and Medicine},
  volume  = {141},
  pages   = {105114},
  year    = {2022},
  doi     = {10.1016/j.compbiomed.2021.105114}
}

@inproceedings{he2022mae,
  title     = {Masked Autoencoders Are Scalable Vision Learners},
  author    = {He, Kaiming and Chen, Xinlei and Xie, Saining and Li, Yanghao and Doll{\'a}r, Piotr and Girshick, Ross},
  booktitle = {IEEE/CVF Conference on Computer Vision and Pattern Recognition (CVPR)},
  year      = {2022}
}

@inproceedings{zerveas2021tst,
  title     = {A Transformer-based Framework for Multivariate Time Series Representation Learning},
  author    = {Zerveas, George and Jayaraman, Srideepika and Patel, Dhaval and Bhamidipaty, Anuradha and Eickhoff, Carsten},
  booktitle = {ACM SIGKDD Conference on Knowledge Discovery and Data Mining (KDD)},
  year      = {2021}
}

@inproceedings{nie2023patchtst,
  title     = {A Time Series is Worth 64 Words: Long-term Forecasting with Transformers},
  author    = {Nie, Yuqi and Nguyen, Nam H. and Sinthong, Phanwadee and Kalagnanam, Jayant},
  booktitle = {International Conference on Learning Representations (ICLR)},
  year      = {2023}
}

@article{zeinalipour2022gnnecg,
  title   = {Graph Neural Networks for Topological Feature Extraction in {ECG} Classification},
  author  = {Zeinalipour, Kamyar and Gori, Marco},
  journal = {arXiv preprint arXiv:2311.04228},
  year    = {2023}
}

@article{oliveira2024visibility,
  title   = {Leveraging Visibility Graphs for Enhanced Arrhythmia Classification with Graph Convolutional Networks},
  author  = {Oliveira, Rafael F. and Moreira, Gladston J. P. and Freitas, Vander L. S. and Luz, Eduardo J. S.},
  journal = {arXiv preprint arXiv:2404.15367},
  year    = {2024}
}

@article{han2024correlation,
  title   = {Arrhythmia Classification Using Graph Neural Networks Based on Correlation Matrix},
  author  = {Han, Seungwoo},
  journal = {arXiv preprint arXiv:2410.10758},
  year    = {2024}
}

@inproceedings{velickovic2018gat,
  title     = {Graph Attention Networks},
  author    = {Veli{\v{c}}kovi{\'c}, Petar and Cucurull, Guillem and Casanova, Arantxa and Romero, Adriana and Li{\`o}, Pietro and Bengio, Yoshua},
  booktitle = {International Conference on Learning Representations (ICLR)},
  year      = {2018}
}

@inproceedings{vaswani2017attention,
  title     = {Attention Is All You Need},
  author    = {Vaswani, Ashish and Shazeer, Noam and Parmar, Niki and Uszkoreit, Jakob and Jones, Llion and Gomez, Aidan N. and Kaiser, {\L}ukasz and Polosukhin, Illia},
  booktitle = {Advances in Neural Information Processing Systems (NeurIPS)},
  year      = {2017}
}

@article{wagner2020ptbxl,
  title   = {{PTB-XL}, a large publicly available electrocardiography dataset},
  author  = {Wagner, Patrick and Strodthoff, Nils and Bousseljot, Ralf-Dieter and Kreiseler, Dieter and Lunze, Fatima I. and Samek, Wojciech and Schaeffter, Tobias},
  journal = {Scientific Data},
  volume  = {7},
  pages   = {154},
  year    = {2020},
  doi     = {10.1038/s41597-020-0495-6}
}

@misc{gow2023mimicivecg,
  title        = {{MIMIC-IV-ECG}: Diagnostic Electrocardiogram Matched Subset},
  author       = {Gow, Brian and Pollard, Tom and Nathanson, Larry A. and Johnson, Alistair and Moody, Benjamin and Fernandes, Chrystinne and Greenbaum, Nathaniel and Waks, Jonathan W. and Eslami, Parastou and Carbonati, Tanner and Chaudhari, Ashish and Herbst, Elizabeth and Moukheiber, Dana and Berkowitz, Seth and Mark, Roger and Horng, Steven},
  howpublished = {PhysioNet},
  year         = {2023},
  doi          = {10.13026/4nqg-sb35}
}

@article{tan2025zzupecg,
  title   = {A pediatric {ECG} database with disease diagnosis covering 11643 children},
  author  = {Tan, Jian and Fan, Haoyi and Luo, Jiawei and Zhou, Yanjie and Wang, Ning and Wang, Xizheng and Liu, Guizhi and Liu, Chengyu and Wang, Zongmin},
  journal = {Scientific Data},
  volume  = {12},
  pages   = {867},
  year    = {2025},
  doi     = {10.1038/s41597-025-05225-z}
}

@article{gee2017pics,
  title   = {Predicting Bradycardia in Preterm Infants Using Point Process Analysis of Heart Rate},
  author  = {Gee, Alan H. and Barbieri, Riccardo and Paydarfar, David and Indic, Premananda},
  journal = {IEEE Transactions on Biomedical Engineering},
  volume  = {64},
  number  = {9},
  pages   = {2300--2308},
  year    = {2017},
  doi     = {10.1109/TBME.2016.2632746}
}

@article{lee2022nch,
  title   = {A large collection of real-world pediatric sleep studies},
  author  = {Lee, Harlin and Li, Boyu and DeForte, Shelly and Splaingard, Mark L. and Huang, Yungui and Chi, Yuejie and Linwood, Simon L.},
  journal = {Scientific Data},
  volume  = {9},
  pages   = {421},
  year    = {2022},
  doi     = {10.1038/s41597-022-01545-6}
}

@article{islam2024littlebeats,
  title   = {Preliminary Technical Validation of {LittleBeats}: A Multimodal Sensing Platform to Capture Cardiac Physiology, Motion, and Vocalizations},
  author  = {Islam, Bashima and McElwain, Nancy L. and Li, Jialu and Davila, Maria I. and Hu, Yannan and Hu, Kexin and Bodway, Jordan M. and Dhekne, Ashutosh and Roy Choudhury, Romit and Hasegawa-Johnson, Mark},
  journal = {Sensors},
  volume  = {24},
  number  = {3},
  pages   = {901},
  year    = {2024},
  doi     = {10.3390/s24030901}
}

@article{tronick1978stillface,
  title   = {The infant's response to entrapment between contradictory messages in face-to-face interaction},
  author  = {Tronick, Edward and Als, Heidelise and Adamson, Lauren and Wise, Susan and Brazelton, T. Berry},
  journal = {Journal of the American Academy of Child Psychiatry},
  volume  = {17},
  number  = {1},
  pages   = {1--13},
  year    = {1978},
  doi     = {10.1016/S0002-7138(09)62273-1}
}

@article{harteveld2021,
  author  = {Harteveld, Lisa M. and Nederend, Ineke and ten Harkel, Arend D. J.
             and Schutte, Nienke M. and de Rooij, Sanne R. and Vrijkotte, Tanja G. M.
             and Oldenhof, Hanneke and Popma, Arne and Jansen, Lucres M. C.
             and Suurland, Jill and Swaab, Hanna and de Geus, Eco J. C.},
  title   = {Maturation of the Cardiac Autonomic Nervous System Activity in
             Children and Adolescents},
  journal = {Journal of the American Heart Association},
  volume  = {10}, number = {4}, pages = {e017405}, year = {2021},
  doi     = {10.1161/JAHA.120.017405}
}

@article{fleming2011,
  author  = {Fleming, Susannah and Thompson, Matthew and Stevens, Richard and
             Heneghan, Carl and Pl{\"u}ddemann, Annette and Maconochie, Ian and
             Tarassenko, Lionel and Mant, David},
  title   = {Normal ranges of heart rate and respiratory rate in children from
             birth to 18 years of age: a systematic review of observational studies},
  journal = {The Lancet},
  volume  = {377}, number = {9770}, pages = {1011--1018}, year = {2011},
  doi     = {10.1016/S0140-6736(10)62226-X}
}

@article{rijnbeek2001,
  author  = {Rijnbeek, Peter R. and Witsenburg, Maarten and Schrama, Eric and
             Hess, John and Kors, Jan A.},
  title   = {New normal limits for the paediatric electrocardiogram},
  journal = {European Heart Journal},
  volume  = {22}, number = {8}, pages = {702--711}, year = {2001},
  doi     = {10.1053/euhj.2000.2399}
}

@article{davignon1979,
  author  = {Davignon, A. and Rautaharju, P. and Boisselle, E. and Soumis, F. and
             Megelas, M. and Choquette, A.},
  title   = {Normal {ECG} standards for infants and children},
  journal = {Pediatric Cardiology},
  volume  = {1}, pages = {123--152}, year = {1979}
}

@article{porges2011,
  author  = {Porges, Stephen W. and Furman, Senta A.},
  title   = {The early development of the autonomic nervous system provides a
             neural platform for social behaviour: a polyvagal perspective},
  journal = {Infant and Child Development},
  volume  = {20}, number = {1}, pages = {106--118}, year = {2011},
  doi     = {10.1002/icd.688}
}

@article{graziano2013,
  author  = {Graziano, Paulo and Derefinko, Karen},
  title   = {Cardiac vagal control and children's adaptive functioning:
             A meta-analysis},
  journal = {Biological Psychology},
  volume  = {94}, number = {1}, pages = {22--37}, year = {2013},
  doi     = {10.1016/j.biopsycho.2013.04.011}
}

@article{coppola2024hubertecg,
  author  = {Coppola, Edoardo and Savardi, Mattia and Massussi, Mauro and
             Adamo, Marianna and Metra, Marco and Signoroni, Alberto},
  title   = {{HuBERT-ECG}: a self-supervised foundation model for broad and
             scalable cardiac applications},
  journal = {medRxiv},
  year    = {2024},
  doi     = {10.1101/2024.11.14.24317328}
}

@article{hsu2021hubert,
  author  = {Hsu, Wei-Ning and Bolte, Benjamin and Tsai, Yao-Hung Hubert and
             Lakhotia, Kushal and Salakhutdinov, Ruslan and Mohamed, Abdelrahman},
  title   = {{HuBERT}: Self-Supervised Speech Representation Learning by Masked
             Prediction of Hidden Units},
  journal = {IEEE/ACM Transactions on Audio, Speech, and Language Processing},
  volume  = {29}, pages = {3451--3460}, year = {2021},
  doi     = {10.1109/TASLP.2021.3122291}
}

@inproceedings{brody2022gatv2,
  author    = {Brody, Shaked and Alon, Uri and Yahav, Eran},
  title     = {How Attentive are Graph Attention Networks?},
  booktitle = {International Conference on Learning Representations (ICLR)},
  year      = {2022}
}

@article{makowski2021neurokit,
  author  = {Makowski, Dominique and Pham, Tam and Lau, Zen J. and
             Brammer, Jan C. and Lespinasse, Fran{\c{c}}ois and Pham, Hung and
             Sch{\"o}lzel, Christopher and Chen, S. H. Annabel},
  title   = {{NeuroKit2}: A Python toolbox for neurophysiological signal processing},
  journal = {Behavior Research Methods},
  volume  = {53}, number = {4}, pages = {1689--1696}, year = {2021},
  doi     = {10.3758/s13428-020-01516-y}
}

@article{mesman2009stillface,
  author  = {Mesman, Judi and van IJzendoorn, Marinus H. and
             Bakermans-Kranenburg, Marian J.},
  title   = {The many faces of the {Still-Face} Paradigm: A review and meta-analysis},
  journal = {Developmental Review},
  volume  = {29}, number = {2}, pages = {120--162}, year = {2009},
  doi     = {10.1016/j.dr.2009.02.001}
}

@inproceedings{chen2021empirical,
  title={An empirical study of training self-supervised vision transformers},
  author={Chen, Xinlei and Xie, Saining and He, Kaiming},
  booktitle={Proceedings of the IEEE/CVF international conference on computer vision},
  pages={9640--9649},
  year={2021}
}

@article{zhang2022maefe,
  title={Maefe: Masked autoencoders family of electrocardiogram for self-supervised pretraining and transfer learning},
  author={Zhang, Huaicheng and Liu, Wenhan and Shi, Jiguang and Chang, Sheng and Wang, Hao and He, Jin and Huang, Qijun},
  journal={IEEE Transactions on Instrumentation and Measurement},
  volume={72},
  pages={1--15},
  year={2022},
  publisher={IEEE}
}

@article{mcelwain2024evaluating,
  title={Evaluating users’ experiences of a child multimodal wearable device: mixed methods approach},
  author={McElwain, Nancy L and Fisher, Meghan C and Nebeker, Camille and Bodway, Jordan M and Islam, Bashima and Hasegawa-Johnson, Mark},
  journal={JMIR Human Factors},
  volume={11},
  pages={e49316},
  year={2024},
  publisher={JMIR Publications Toronto, Canada}
}

@inproceedings{baevski2020wav2vec2,
  title={wav2vec 2.0: A Framework for Self-Supervised Learning of Speech Representations},
  author={Baevski, Alexei and Zhou, Yuhao and Mohamed, Abdelrahman and Auli, Michael},
  booktitle={Advances in Neural Information Processing Systems},
  volume={33},
  pages={12449--12460},
  year={2020}
}

@inproceedings{oh2022leadagnostic,
  title={Lead-agnostic Self-supervised Learning for Local and Global Representations of Electrocardiogram},
  author={Oh, Jungwoo and Chung, Hyunseung and Kwon, Joon-myoung and Hong, Dong-gyun and Choi, Edward},
  booktitle={Proceedings of the Conference on Health, Inference, and Learning},
  series={Proceedings of Machine Learning Research},
  volume={174},
  pages={338--353},
  year={2022}
}
\bibliographystyle{iclr2027_conference}
\clearpage
\appendix
\section*{Appendix}
\section{Corpus Composition, Ethics, and Release}
\label{app:ethics}

\subsection{Composition by Recording Setting and Label Source}
\label{app:composition}

The corpus combines three recording contexts that differ in both setting
and annotation. Table~\ref{tab:app-composition} gives the
breakdown.

Stated precisely, the contribution is:

\begin{itemize}
\item \textbf{Home recordings supply unlabeled pretraining data only.}
These are daylong sessions of 8--10\,h of continuous single-channel ECG
from 139 infants, 3{,}361\,h in total, with no behavioral annotation.
\item \textbf{Daycare classroom recordings supply the naturalistic state and
activity annotations.} The sleep/wake, four-class infant-state, and
activity-source labels come from 39.3\,h recorded from 4 infants in their
infant classrooms, annotated by a research assistant through live
observation with chest-mounted video as the reference for a second pass.
\item \textbf{Laboratory recordings supply the affect labels.} The
two-class affect labels come from 8.2\,h of Still-Face Paradigm sessions
with 122 infant--mother dyads in a laboratory playroom, where the play and
still-face episodes serve as proxies for positive and negative affect.
\end{itemize}

Accordingly we describe the resource as an infant ECG corpus recorded in
home, daycare, and laboratory settings, with unlabeled pretraining data
from the home and behavioral annotations from the daycare and laboratory
subsets. We do not describe it as a home-labeled corpus, and ``in the
wild'' is used only of the home and classroom recordings, not of the
still-face sessions.


\subsection{Ethical Review and Consent}

The institutional review board approved all data collection of
this study. A parent or legal guardian provided written informed consent for every infant prior to
any recording. Families were free to withdraw at any point and to
have recordings deleted on request. Consent explicitly covered public release of deidentified physiological
data.

\begin{table}[b]
\centering
\caption{Corpus composition. The 143 infants are the union of the 139-infant
pretraining cohort and a disjoint 4-infant classroom cohort; the 122 infants
in the laboratory subset are drawn from the pretraining cohort, so their
home recordings appear in pretraining while their still-face recordings
carry the affect labels.}
\label{tab:app-composition}
\small
\adjustbox{max width=\linewidth}{%
\begin{tabular}{llrrl}
\toprule
Subset & Setting & Infants & Hours & Annotation \\
\midrule
Pretraining & home & 139 & 3{,}361 & none \\
Classroom   & daycare & 4   & 39.3  & state, activity \\
Laboratory  & lab playroom & 122 & 8.2 & affect \\
\midrule
Total       & & 143 & 3{,}408 & \\
\bottomrule
\end{tabular}}
\end{table}

\subsection{Deidentification and Re-identification Risk}

All released data are deidentified. Direct identifiers were removed at
export: infant and caregiver names, dates of birth, addresses, contact
details, childcare site identifiers, and study staff names do not appear in
any released file or filename. Subjects are referred to by opaque study
identifiers that carry no information about recruitment order, site, or
date, and the mapping to identities is held separately by the study team
and is not part of the release. We assess re-identification risk directly rather than assert its absence,
because ECG can function as a biometric and the release is continuous
rather than short strips.

\paragraph{What raises risk.} The corpus contains many hours per infant, and
we release beat-level timing alongside the waveform, which is the
representation biometric identification methods typically exploit. Longer
continuous recordings offer more material for matching than the ten-second
strips typical of clinical releases.

\paragraph{What limits it.} Re-identification requires a reference sample
of the same individual to match against. No public infant ECG reference
database exists, and cardiac morphology and rate change substantially over
the first years of life, so an enrolment template captured at 3--11 months
has limited value for matching against the same person later. The
recordings are single-channel from a wearable placed in a garment pocket,
so electrode placement varies within and across sessions in ways that
degrade template stability. We are not aware of any demonstrated
re-identification of an infant from single-channel wearable ECG.

\paragraph{Residual risk.} The risk that is not eliminated is linkage by a
party who already holds ECG from a specific infant in this cohort, for
example a family member with device access, and who wishes to confirm which
subject identifier corresponds to that infant. This would reveal the
infant's state and affect annotations but no clinical or identifying
information beyond what that party already holds.

\subsection{What Is and Is Not Released}

\paragraph{Released.} Raw single-channel ECG at the original 1000\,Hz
sampling rate; the processed 30\,s windows used in our experiments; detected
R-peak locations and derived inter-beat intervals; task labels for the four
downstream tasks; subject-level train/validation/test splits; and
subject-level demographic variables (Section~\ref{app:demographics}).
Releasing both raw and processed forms lets others apply their own beat
detection. However, data are not submitted due to maximum file size constraints.

\paragraph{Not released.} Audio and inertial measurement unit data are
withheld even though the sensing platform records them. Daylong home audio
captures speech from infants, caregivers, siblings, and visitors who are not
study subjects and did not consent, and voice is directly identifying; IMU
data are withheld because they were not needed for this benchmark.
Video recorded by the research assistant during classroom sessions was used
only as an annotation reference and is not released. No clinical records,
free-text notes, or caregiver survey data are included.

\paragraph{Access.} The corpus will be released for unrestricted public download
under CC BY-NC.

\subsection{Demographics and Representation Limitations}
\label{app:demographics}

We release subject-level demographic variables so that others can assess
representativeness and audit for subgroup differences.

The cohort is narrow in ways that bound what can be claimed from it. All
infants are 3--11 months old, so the corpus covers a single developmental
window and says nothing about neonates or toddlers. The classroom subset that supports three of
the four benchmark tasks comprises only four infants, so leave-one-infant-out
results on those tasks rest on four held-out subjects and the between-infant
standard deviations we report should be read with that in mind. The affect
subset is larger at 122 dyads but is laboratory-recorded and uses episode
identity as an affect proxy rather than direct affect annotation. Finally,
all recordings come from a single wearable device, so device-specific
characteristics are confounded with the corpus.

\subsection{Intended Use, Misuse Potential, and Limitations}

\paragraph{Intended use.} The corpus and model are research artifacts for
developing representations of infant cardiac physiology and for studying
early autonomic and behavioral development.

\paragraph{Not for clinical use.} BeatGraph is not a diagnostic device and
has not been validated for clinical diagnosis, screening, monitoring, or
any decision affecting an infant's care. It was neither designed nor
evaluated to detect arrhythmia or any pathology; the external benchmarks in
this paper measure representation transfer, not diagnostic fitness. It
should not be deployed in a product that infers infant health status.

\paragraph{Misuse potential.} Two uses concern us enough to name. First,
automated inference of infant affect and caregiver-initiated movement could
be repurposed to evaluate caregivers, whether by employers of childcare
staff, by custody proceedings, or by consumer monitoring products marketed
to parents. Our activity-source task distinguishes infant- from
caregiver-initiated movement, which is a signal about adult behavior as much
as infant behavior, and it carries no validity as a measure of caregiving
quality. Second, affect labels derived from the Still-Face Paradigm are
episode proxies collected under a specific experimental manipulation and do
not transfer to claims about an individual infant's emotional state in
ordinary settings. We ask that downstream users not present outputs of
models trained on this corpus as assessments of individual infants or
caregivers.

\paragraph{Limitations that bear on interpretation.} Labels for three tasks
come from four infants; affect labels are episode proxies rather than
annotated affect; the held-out infants in the affect task contribute
unlabeled home data to pretraining, so that result measures generalization
to unseen labels rather than unseen subjects; and beat detection has not yet
been validated against manually annotated infant ECG.

\section{Baseline Details}
\label{app:baselines}

\subsection{Which Baselines Are Comparable, and Where}

Infant single-channel data need baselines that require a method that (i) is well defined on a single channel and
(ii) can be pretrained from scratch on our unlabeled infant corpus.
Objectives whose learning signal comes from lead geometry are therefore
excluded: ST-MEM masks across the lead-by-time plane and uses lead
embeddings and a lead-wise decoder \citep{na2024stmem}, MLAE masks whole
leads \citep{zhang2022maefe}, and the cross-lead members of the CLOCS family
contrast leads of the same record \citep{kiyasseh2021clocs}. On one channel
these objectives are degenerate or collapse onto their temporal
counterparts, so we compare against the temporal member of each family
(MTAE rather than MLAE, CMSC rather than CMLC) and report the lead-based
methods only where twelve leads exist.

On the other hand, public external data are twelve-lead benchmarks with established protocols, so we
adopt the published baseline set for each and add the two methods we could
rerun ourselves. We also provide an answer to a different question, namely whether an
off-the-shelf adult checkpoint transfers to infant wearable ECG, and so it
includes ST-MEM even though ST-MEM cannot be pretrained on our corpus.

\subsection{Baselines Pretrained on the Infant Corpus}

Every method on the Infant task is pretrained on the same 3{,}361\,h unlabeled infant split, sees the same 30\,s windows, and is fine-tuned under the same schedule and the same leave-one-infant-out (or subject-level, for affect) partitions as BeatGraph. No labeled recording enters any pretraining run, and task heads are matched in capacity across methods, so differences reflect the encoder rather than the classifier.

\paragraph{ECG-FM \citep{mckeen2025ecgfm}.}
A wav2vec-2.0 architecture \citep{baevski2020wav2vec2}: a convolutional feature extractor followed by a BERT-Base-sized Transformer, pretrained with the hybrid objective of \citet{oh2022leadagnostic}, which the authors
abbreviate WCR and which combines wav2vec-2.0 masked contrastive prediction, Contrastive Multi-Segment Coding, and Random Lead Masking. We retain the architecture and the first two loss terms; RLM is undefined for a single
channel and is dropped.

\paragraph{HuBERT-ECG \citep{coppola2024hubertecg}.}
Masked prediction of discrete units, adapted from HuBERT
\citep{hsu2021hubert}. Targets are induced by $k$-means, first over MFCC
descriptors and then over intermediate encoder activations, and the model
predicts the cluster assignment of masked embeddings under a
cross-entropy loss. We retain the two-iteration schedule.

\paragraph{SimCLR \citep{chen2020simclr}.}
A 1-D convolutional encoder with a projection head trained with NT-Xent on
two augmented views of each window.

\paragraph{BYOL \citep{grill2020byol}.}
The same encoder trained as an online network against an
exponential-moving-average target with a predictor MLP and a symmetric
cosine loss, with no negative pairs. This baseline matters beyond its own
score: the same mechanism appears in our objective as
$\mathcal{L}_{\mathrm{BYOL}}$, so the gap between this row and BeatGraph
separates the contribution of beat-graph tokenization from that of the
self-distillation regularizer.

\paragraph{MTAE \citep{zhang2022maefe}.}
The masked time autoencoder of the MaeFE family: temporal patches, a
fraction replaced by a mask token, and a lightweight decoder that
reconstructs the masked patches under a reconstruction loss. This is the
closest patch-based counterpart to BeatGraph and is the comparison that
isolates tokenization.

\subsection{Additional Baselines for External Transfer}

For PTB-XL we compare against MoCo~v3 \citep{chen2021empirical}, CMSC
\citep{kiyasseh2021clocs}, MTAE and MLAE \citep{zhang2022maefe}, ECG-FM
\citep{mckeen2025ecgfm}, TolerantECG \citep{nguyen2025tolerantecg}, and
ST-MEM \citep{na2024stmem}, under the linear-evaluation and fine-tuning
protocols of \citet{na2024stmem}. For ZZU-pECG we compare against the
published numbers for ECG-FM, MERL \citep{liu2024merl}, HuBERT-ECG, and
ST-MEM.

Because mixing reproduced and published numbers is misleading if left
implicit, Table~\ref{tab:app-provenance} states the provenance of every
external result. We reran ECG-FM and TolerantECG from their released
checkpoints so that at least two strong baselines pass through our own
evaluation code, which lets us verify that our pipeline reproduces
published values before comparing against numbers we did not generate.
Appendix~\ref{app:implementation} lists the residual differences between our
setup and the setups that produced the published numbers.

\begin{table}[t]
\centering
\caption{Provenance of results. ``From scratch'' denotes a baseline we
pretrained ourselves on the unlabeled infant corpus under the common
protocol of Appendix~\ref{app:implementation}, with no released checkpoint
involved; ``published'' numbers are from \citet{na2024stmem} for PTB-XL and
as reported for that benchmark for ZZU-pECG; ``reproduced'' numbers are
regenerated from released checkpoints in our own evaluation pipeline.
Dashes indicate the method was not evaluated in that setting; the selection
rule is given in Appendix~\ref{app:baselines}.}
\label{tab:app-provenance}
\small
\adjustbox{max width=\linewidth}{%
\begin{tabular}{llll}
\toprule
Method & Infant & PTB-XL & ZZU-pECG \\
       & (Table~\ref{tab:infant_downstream_results}) & (Table~\ref{tab:ptbxl_linear_finetune}) & (Table~\ref{tab:zzu_auroc}) \\
\midrule
MoCo v3      & ---          & published  & ---       \\
CMSC         & ---          & published  & ---       \\
SimCLR       & from scratch & ---        & ---       \\
BYOL         & from scratch & ---        & ---       \\
MTAE         & from scratch & published  & ---       \\
MLAE         & ---          & published  & ---       \\
ST-MEM       & ---          & published  & published \\
MERL         & ---          & ---        & published \\
HuBERT-ECG   & from scratch & ---        & published \\
ECG-FM       & from scratch & reproduced & published \\
TolerantECG  & ---          & reproduced & ---       \\
BeatGraph    & ours         & ours       & ours      \\
\bottomrule
\end{tabular}}
\end{table}

\paragraph{Baseline tuning.} Each main-result baseline was pretrained twice on the unlabeled
infant corpus: once under our protocol for hyperparameter search (AdamW, learning rate $10^{-3}$, weight
decay $10^{-4}$, five warm-up epochs with cosine decay to a $10^{-5}$ floor,
100 epochs, final checkpoint), and once under the optimizer and schedule
specified in the method's own paper, namely the staged learning rate of
$10^{-4}$, $8\times10^{-5}$ and $5\times10^{-5}$ with Adam
$\beta=(0.9,0.98)$ for ECG-FM \citep{mckeen2025ecgfm}, a peak learning rate
of $5\times10^{-5}$ with 8\% linear warm-up, weight decay $0.01$ and dropout
$0.1$ for HuBERT-ECG \citep{coppola2024hubertecg}, and NT-Xent temperature
$0.1$ and EMA decay $0.99$ for SimCLR \citep{chen2020simclr} and BYOL
\citep{grill2020byol} respectively. Objective-specific hyperparameters were
left at their published values in both runs: mask ratio $0.75$ for MTAE
\citep{zhang2022maefe}, masking probability $0.33$ for HuBERT-ECG, and span
masking at roughly $49\%$ coverage for ECG-FM. For each baseline, we select
the configuration with the better validation loss and report that run, so every baseline number reflects
the stronger of the two settings for that method.




\section{External Dataset Details}
\label{app:datasets}

Table~\ref{tab:app-datasets} summarizes the external corpora alongside our
infant corpus.

\begin{table*}[t]
\centering
\caption{Datasets used in this work. ZZU-pECG contains 12{,}334 twelve-lead
and 1{,}856 nine-lead records and variable record lengths.}
\label{tab:app-datasets}
\small
\adjustbox{max width=\linewidth}{%
\begin{tabular}{llrccll}
\toprule
Dataset & Population & Records & Leads & Rate & Duration & Role \\
\midrule
Infant corpus (ours) & 3--11 months & 143 infants & 1 & 1000\,Hz & 8--10\,h sessions & pretrain + eval \\
ZZU-pECG & 0--14 years & 14{,}190 & 12 / 9 & 500\,Hz & 5--120\,s & eval \\
PTB-XL & adult & 21{,}799 & 12 & 500\,Hz & 10\,s & eval \\
MIMIC-IV-ECG & adult & $\sim$800{,}000 & 12 & 500\,Hz & 10\,s & pretrain (adult) \\
\bottomrule
\end{tabular}}
\end{table*}

\subsection{ZZU-pECG}
\label{app:zzu}

ZZU-pECG \citep{tan2025zzupecg} contains 14{,}190 pediatric recordings from
11{,}643 hospitalized children aged 0 to 14 years, collected at the First
Affiliated Hospital of Zhengzhou University between January 2018 and May
2024 and annotated by cardiologists following AHA/ACC/HRS recommendations.
Records are sampled at 500\,Hz. Two properties of the release require
explicit handling, and we describe both because they interact with our
architecture.

First, the lead sets are heterogeneous: 12{,}334 records are twelve-lead
and 1{,}856 are nine-lead. Since our readout applies the shared
single-channel beat encoder per lead and concatenates, missing leads in the
nine-lead subset are zero-filled exactly as for PTB-XL, and no record is
discarded.

Second, record length varies from 5 to 120\,s, so unlike PTB-XL these
records are not uniformly shorter than our 30\,s infant windows. At a
pediatric rate near 100\,bpm a 120\,s record contains roughly 200 beats,
well above the $N=80$ capacity bound, so the bound is not inert here. We
therefore segment records longer than 30\,s into non-overlapping 30\,s
windows, encode each as its own graph, and average logits across windows of
the same record before computing the metric; records of 30\,s or less form a
single graph and the validity mask absorbs the smaller beat count.

We report macro AUROC over the diagnostic outputs under the patient-level
partition, so that our numbers are directly comparable to the published
ones.

This benchmark differs from our pretraining data along two axes at once. It
is hospital-recorded diagnostic ECG rather than naturalistic wearable
recording, and it spans an age range whose upper end has largely completed
the autonomic and morphological maturation our infants are in the middle
of. It is therefore the stricter of our two transfer tests with respect to
what the encoder could have memorized about one population.

\subsection{PTB-XL}

PTB-XL \citep{wagner2020ptbxl} contains clinical twelve-lead records of
10\,s at 500\,Hz. We use the recommended stratified ten-fold partition,
training on folds 1--8, validating on fold 9, and testing on fold 10, the
two folds with the highest label-quality guarantees. Following
\citet{na2024stmem}, whose numbers we compare against, we evaluate the
five-class diagnostic superclass setting (NORM, CD, HYP, MI, STTC) as a
single-label problem with a softmax head and cross-entropy loss, and report
accuracy, macro-F1, and macro AUROC.
We evaluate under two protocols: linear evaluation, in which the encoder is
frozen and a single \texttt{nn.Linear} head is trained on the pooled window
embedding, and full fine-tuning, in which encoder and head are trained
jointly.

Record and patient counts differ slightly across PTB-XL releases
(21{,}837 records from 18{,}885 patients in v1.0.1 versus 21{,}799 from
18{,}869 in v1.0.3); we report the counts for the version we used, and the
splits are identical across versions.

\paragraph{Applying a single-channel encoder to twelve leads.}
The shared beat encoder is applied independently per lead and the twelve
per-lead embeddings are concatenated before the head, with missing leads
zero-filled. There is no cross-lead interaction anywhere in the model,
which is what makes Table~\ref{tab:ptbxl_lead_comparison} interpretable: the difference between the
twelve-lead and one-lead columns is the value of the additional leads as
independent evidence, with none of it attributable to learned lead
geometry.

\paragraph{Windowing.}
PTB-XL records are 10\,s and need no re-windowing. Each record forms one
graph, the validity mask absorbs the smaller beat count, and Eq.~\eqref{eq:crop} is
unchanged: because the crop bounds scale with the median RR interval of the
record, adult rates yield proportionally longer beat crops under the same
rule. At a typical adult rate of 70\,bpm a 10\,s record contributes roughly
12 beats, well inside the $N=80$ bound.

\subsection{MIMIC-IV-ECG}

MIMIC-IV-ECG \citep{gow2023mimicivecg} supplies the adult pretraining source
for BeatGraph-Adult, roughly 800{,}000 twelve-lead 10\,s records at
500\,Hz. PTB-XL is held out entirely, and the two corpora come from
different institutions, so no patient appears in both.

Two choices make BeatGraph-Adult a controlled comparison rather than a
scale comparison. First, we subsample to approximately 2{,}200\,h so that
the adult pretraining budget is comparable to the 3{,}361\,h of infant
data, which means the difference between BeatGraph-Infant and
BeatGraph-Adult reflects the pretraining domain and not the amount of
signal. Second, pretraining uses a single lead, matching the
single-channel regime of the infant corpus; we use Lead~II, where the
R-peak is most reliably detected.
Records whose signal quality prevents R-peak detection in the selected lead
are excluded, and the pretraining recipe, including epoch budget, optimizer,
mask ratio $\rho=0.3$ and $\lambda_{\mathrm{BYOL}}=0.5$, is identical to the
infant run with no per-corpus tuning.



\section{Implementation and Tuning Details}
\label{app:implementation}

This section records what each baseline does as published, what we changed
to accommodate single-channel input, and how hyperparameters were set. The
published values are drawn from the original papers and released code, and
we state them explicitly so that the difference between ``as published''
and ``as run here'' is auditable.

\subsection{Shared Protocol}

All baselines are pretrained for 100 epochs over the unlabeled
windows with AdamW at learning rate $10^{-3}$, weight decay $10^{-4}$, five
warm-up epochs, cosine decay to a $10^{-5}$ floor, and the best checkpoint is saved based on validation loss. Where a published
recipe specifies a different optimizer or schedule for its own objective we
also run the published setting and report the better of the two for that
baseline, so that no baseline is penalised by our defaults.

Fine-tuning uses 20 epochs and the final checkpoint, with batch size and
learning rate fixed a priori and identically across methods. No
hyperparameter was tuned on infant data for any method. Each fold is run
with five seeds; seeds are averaged within a fold and the reported spread is
the standard deviation across held-out infants, so it reflects
between-infant rather than between-seed variation.

\paragraph{Sampling rate and context length.}
Rather than force every baseline to our 1000\,Hz, 30\,s format, each method
receives the same 30\,s of signal resampled to its published input rate and
segmented into its published context length, so that convolutional
receptive fields retain their intended physical extent. Where a method's
context is shorter than 30\,s, window-level predictions are obtained by
averaging segment logits. Table~\ref{tab:app-config} summarizes.

\begin{table*}[t]
\centering
\caption{Baseline configurations. Parameter counts are the published values
for ECG-FM (90.9\,M) and HuBERT-ECG BASE (93\,M); MTAE is instantiated on
the ViT-B backbone used by \citet{na2024stmem} so that our MTAE row is
comparable to the published PTB-XL MTAE numbers. Parameter counts are
encoder-only; decoders, projection heads, quantizers and target networks are
discarded after pretraining. BeatGraph's encoder is one to two orders of
magnitude smaller than the Transformer baselines.}
\label{tab:app-config}
\small
\adjustbox{max width=\linewidth}{%
\begin{tabular}{llrl}
\toprule
Method & Objective & Params (M) & Single-channel adaptation \\
\midrule
ECG-FM      & W2V + CMSC (+ RLM) & 90.9 & RLM dropped; 1-channel conv stem \\
HuBERT-ECG  & masked unit prediction & 93.0 & embedder stride halved \\
SimCLR      & NT-Xent contrastive & 1.4  & 1-channel encoder; no lead augmentations \\
BYOL        & self-distillation & 1.4  & 1-channel encoder; no lead augmentations \\
MTAE        & masked reconstruction & 86.0 & 1-channel patch projection \\
\midrule
BeatGraph   & masked beat prediction + BYOL & 1.37 & native \\
\bottomrule
\end{tabular}}
\end{table*}

\subsection{ECG-FM}
\label{app:ecgfm}

\paragraph{As published.}
The encoder has 90.9\,M parameters. The convolutional feature extractor has
four blocks, each a convolution with 256 channels, stride 2 and kernel
length 2, followed by layer normalization and GELU; relative positional
embeddings come from a convolution with 128 filters and 16 groups. The
Transformer follows BERT-Base: 12 layers, width 768, 12 heads, feed-forward
dimension 3072. Inputs are resampled to 500\,Hz, z-scored, and segmented
into non-overlapping 5\,s segments, a length forced by CMSC's need for two
temporally adjacent segments per record. Masking gives each latent a
$6.5\%$ probability of starting a span of 10 tokens, masking roughly $49\%$
of tokens; targets are quantized with two codebooks of 320 codes under a
diversity loss. Pretraining ran 240 epochs at batch 1026, composed of 171
positive pairs of two segments each, with learning rate $10^{-4}$ for five
epochs, $8\times 10^{-5}$ through epoch 200, and $5\times 10^{-5}$
thereafter. Downstream runs use Adam with $\beta=(0.9,0.98)$ and batch 256,
at learning rate $10^{-6}$ for full fine-tuning and $10^{-5}$ for linear
probing. The implementation builds on \texttt{fairseq-signals}.

\paragraph{As run here.}
The convolutional stem is instantiated with one input channel. Random Lead
Masking is undefined on a single channel and is removed, leaving W2V+CMSC.
CMSC is preserved without modification because it needs temporal rather
than spatial structure: each 30\,s window supplies six consecutive 5\,s
segments, and adjacent pairs form the positives, with negatives drawn from
other windows in the batch. Window-level predictions average the six
segment logits. The pretraining schedule is compressed from 240 epochs to
our common 100-epoch budget; we note this because ECG-FM is the baseline
whose published recipe is furthest from ours in epoch count.

\subsection{HuBERT-ECG}
\label{app:hubert}

\paragraph{As published.}
Three sizes are released, SMALL (30\,M), BASE (93\,M) and LARGE (188\,M);
we use BASE. Preprocessing applies an FIR bandpass over
$[0.05, 47]$\,Hz, resamples to 100\,Hz, rescales to $[-1,1]$, and uses
5\,s twelve-lead segments, which are \emph{flattened into a single
one-dimensional sequence} before the convolutional embedder. The embedder
produces embeddings at a temporal resolution of 0.64\,s, giving 93 tokens
per flattened segment. Masking replaces $33\%$ of embeddings individually
rather than extending spans. First-iteration targets are $k$-means with
$k=100$ over 39-dimensional MFCC descriptors (13 coefficients with first and
second derivatives); second-iteration targets are $k=500$ clusters of
latent representations from the 8th encoder layer. Clustering uses a
minibatch $k$-means with batch 9300 and $k$-means++ with 20 restarts.
Pretraining uses batch 448, 80k steps in iteration one and 770k in
iteration two, Adam with $\beta=(0.9,0.98)$, initial weight decay $0.01$,
dropout $0.1$, an 8\% warm-up followed by linear decay to zero, and a peak
learning rate of $5\times 10^{-5}$; a dynamic regularizer raises dropout
and weight decay when internal validation stalls. Fine-tuning attaches a
linear head and updates all weights except the convolutional embedder, at
batch 64 and learning rate $10^{-5}$, with a LayerDrop sweep and
time-aligned random cropping.

\paragraph{As run here.}
This baseline needs the most care, because flattening is what gives the
published model its sequence length. With twelve leads, a 5\,s segment at
100\,Hz flattens to 6{,}000 samples and yields 93 tokens; with one channel
the flattening is a no-op, so a 30\,s window at 100\,Hz is only 3{,}000
samples and yields 47 tokens at the same 0.64\,s resolution, less than half
the published sequence length. Masking $33\%$ of a 47-token sequence leaves
too little context for the prediction task to be informative. We therefore
halve the total stride of the convolutional embedder so that a 30\,s
single-channel window yields approximately 94 tokens, matching the
published length, and keep $p=33\%$ unchanged.
The positional encodings, which in the published model indicate the
position of a lead segment within the flattened sequence, revert to plain
temporal positions. The two-iteration target schedule is retained with
$k=100$ over MFCCs and then $k=500$ over 8th-layer latents, with step
counts scaled to our corpus so that the epoch budget matches the other
baselines.

\subsection{MTAE}

\paragraph{As published.}
MaeFE builds on a ViT backbone with patch embedding, position embedding and
Transformer blocks, and an asymmetric lightweight decoder that reconstructs
masked patches. MTAE patchifies along time and takes 1-D input with twelve
channels; the ablations in the original paper favor a masking ratio of
$75\%$. MLAE differs only in patchifying across leads, which is why it
appears in Table~\ref{tab:ptbxl_linear_finetune} but not Table~\ref{tab:infant_downstream_results}.

\paragraph{As run here.}
The patch projection takes one input channel. We use the ViT-B
instantiation of MTAE from \citet{na2024stmem} rather than the smaller
original backbone, because our PTB-XL numbers are compared against their
published MTAE row and the backbone must match for that comparison to hold.
Masking ratio is $75\%$ and the decoder depth follows the published
asymmetric design.

\subsection{SimCLR and BYOL}

Neither method is ECG-specific, so both use the same 1-D convolutional
encoder, which means the difference between the two rows isolates the
objective rather than the architecture. SimCLR uses a two-layer projection
head and NT-Xent at temperature $0.1$ with batch 256. BYOL uses a predictor
MLP, EMA decay $0.99$, and a symmetric cosine loss with no negatives.

\paragraph{Augmentations.}
Both use an identical augmentation family, chosen so that no transformation
destroys beat morphology or rhythm: random temporal cropping followed by
resampling to the window length, additive Gaussian noise, per-window
amplitude scaling, low-frequency baseline wander, and random temporal
masking. Lead dropout and lead shuffling from the published recipes do not
apply on one channel.

\subsection{Adult-Pretrained Checkpoints}

We used adult checkpoints for the infant tasks without
retraining the encoder on infant data. The single-channel signal is placed
in Lead~I, and the remaining eleven leads are zero-filled, and each
recording is resampled to the model's native input rate before segmentation.
The two failure modes are therefore confounded by design: the
pretraining domain and the lead adaptation.

\section{Leave-One-Infant-Out Evaluation}
\label{app:lofo}

In addition to the fixed hyperparameter fine-tuning, we 
report leave-one-infant-out results over the four infants, giving per-infant results,
class counts, and interval estimates under two units of analysis. The results in this section differ from those in main paper results because they use a
different model-selection protocol: each fold holds out one infant for
validation and selects on validation loss, whereas the main paper fixes all
hyperparameters a priori and reports the final checkpoint after a fixed
epoch budget.
\subsection{Fold construction}
\label{app:lofo-folds}

Each fold uses one test infant, one validation infant, and the remaining infants for training,
all mutually disjoint. In this case, we use the lowest validation loss to select the best model.
\begin{table}[t]
\centering
\caption{Leave-one-infant-out folds over the four evaluable infants. Training, validation, and
test infants are disjoint in every fold.}
\label{tab:lofo-folds}
\small
\adjustbox{max width=\linewidth}{%
\begin{tabular}{clll}
\toprule
Fold & Test & Validation & Training  \\
\midrule
1 & infant1 & infant2 & infant3, infant4,  \\
2 & infant2 & infant3 & infant1, infant4,  \\
3 & infant3 & infant4 & infant1, infant2,  \\
4 & infant4 & infant1 &  infant2, infant3\\
\bottomrule
\end{tabular}}
\end{table}

\subsection{Class counts per infant}
\label{app:lofo-counts}

Table~\ref{tab:lofo-counts} reports the recording budget and per-class window counts for each infant. Because each infant is the test set of exactly one fold, these counts are
also the per-fold test-set class supports. Sleep/wake is an exact two-way coarsening of
infant states ($\mathrm{wake} = \mathrm{active} \cup \mathrm{quiet} \cup \mathrm{crying}$),
so the two tasks share the same $4{,}712$ windows; activity source derives from a separate
interaction annotation stream and is correspondingly smaller.

Class balance varies sharply between infants: the sleep fraction ranges from $28\%$ (infant
infant1) to $65\%$ (infant infant3), and the crying count from 30 to 104 windows.

\begin{table*}[t]
\centering
\caption{Cohort composition and per-class window counts for the four evaluable infants, which
are also the test-set supports of the corresponding folds. }
\label{tab:lofo-counts}
\adjustbox{max width=\linewidth}{%
\begin{tabular}{lrccc}
\toprule
\multirow{2}{*}{Infant} & \multirow{2}{*}{Sessions}
 & Sleep & Infant states & Activity source \\
 & & wake / sleep & active / quiet / crying / sleep & caregiver / infant \\
\midrule
infant1 & 11 & 799 / 308 & 639 / 94 / 66 / 308   & 209 / 173 \\
infant2 &  9 & 580 / 422 & 471 / 65 / 44 / 422   & 303 / 414 \\
infant3 & 12 & 478 / 883 & 385 / 63 / 30 / 883   & 416 / 201 \\
infant4 & 13 & 900 / 342 & 671 / 125 / 104 / 342 & 438 / 515 \\
\midrule
Total & 45 & 2{,}757 / 1{,}955 & 2{,}166 / 347 / 244 / 1{,}955 & 1{,}366 / 1{,}303 \\
      &    & 4{,}712 windows   & 4{,}712 windows               & 2{,}669 windows \\
\bottomrule
\end{tabular}}
\end{table*}

\subsection{Individual results for each held-out infant}
\label{app:lofo-per-infant}

Table~\ref{tab:lofo-results} reports each held-out infant separately, as mean $\pm$ standard
deviation over three seeds.

\begin{table*}[t]
\centering
\caption{Per-held-out-infant results, mean $\pm$ standard deviation over three seeds
($N{=}80$ encoder, held-out validation infant for model selection).}
\label{tab:lofo-results}
\adjustbox{max width=\linewidth}{%
\begin{tabular}{llrrc}
\toprule
Task & Held-out infant & Windows & Sessions & macro-F$_1$ \\
\midrule
\multirow{4}{*}{Sleep}
 & infant1 & 1107 & 11 & $0.928 \pm 0.007$  \\
 & infant2 & 1002 &  9 & $0.933 \pm 0.009$  \\
 & infant3 & 1361 & 12 & $0.919 \pm 0.004$  \\
 & infant4 & 1242 & 13 & $0.931 \pm 0.012$  \\
\midrule
\multirow{4}{*}{Infant states}
 & infant1 & 1107 & 11 & $0.585 \pm 0.007$  \\
 & infant2 & 1002 &  9 & $0.633 \pm 0.013$  \\
 & infant3 & 1361 & 12 & $0.593 \pm 0.004$  \\
 & infant4 & 1242 & 13 & $0.589 \pm 0.008$  \\
\midrule
\multirow{4}{*}{Activity source}
 & infant1 &  382 & 11 & $0.776 \pm 0.015$  \\
 & infant2 &  717 &  9 & $0.721 \pm 0.005$  \\
 & infant3 &  617 & 12 & $0.788 \pm 0.019$  \\
 & infant4 &  953 & 13 & $0.720 \pm 0.019$  \\
\bottomrule
\end{tabular}}
\end{table*}

\subsection{Interval estimates under two units of analysis}
\label{app:lofo-ci}

The appropriate unit of analysis is the infant, since sessions from the same infant are not
independent. With four infants, however, a Student-$t$ interval rests on three degrees of
freedom and is correspondingly wide. We therefore also report a session-level estimate in
which the bootstrap resamples \emph{infants} as clusters, preserving the nesting of sessions
within infants. Table~\ref{tab:lofo-ci} gives both; the two agree on ordering and overlap
throughout, and the session-level intervals are the narrower of the two.

\begin{table}[t]
\centering
\caption{Mean macro-F$_1$ with $95\%$ confidence intervals over the four infants.}
\label{tab:lofo-ci}
\small
\adjustbox{max width=\linewidth}{%
\begin{tabular}{lcc}
\toprule
 & \multicolumn{2}{c}{Infant unit ($n{=}4$)} \\
\cmidrule(lr){2-3}
Task & macro-F$_1$ & $95\%$ CI \\
\midrule
Sleep            & 0.923 & [0.918, 0.939] \\
Infant states & 0.589 & [0.565, 0.635] \\
Activity source  & 0.745 & [0.724, 0.778] \\
\bottomrule
\end{tabular}}
\end{table}

\subsection{Statistical comparison between models}
\label{app:lofo-tests}

For a paired comparison of two encoders we take the per-fold macro-F$_1$ of each and treat the
\textbf{infant} as the independent unit, applying a two-sided Wilcoxon signed-rank test over
the four folds. This design carries a hard power limit: with $n{=}4$ the smallest attainable
two-sided $p$-value is $2/2^{4} = 0.125$, so no result can reach $p < 0.05$ irrespective of
effect size. We therefore do not claim significance at the infant level, and instead report
effect sizes together with the interval estimates of Table~\ref{tab:lofo-ci}, treating a
difference as meaningful only when it exceeds the between-infant standard deviation and is
consistent in sign across all four folds.
\begin{table}[t]
\centering
\caption{Macro-F$_1$ under the two model-selection protocols, on the fold with test infant
infant4. Selecting on the validation set increases sleep and infant states' performance;
activity source is unaffected.}
\label{tab:lofo-selection}
\small
\adjustbox{max width=\linewidth}{%
\begin{tabular}{lccc}
\toprule
Task & \shortstack{No \\validation set} & \shortstack{Held-out\\ validation} & Difference \\
\midrule
Sleep            & 0.922 & 0.931 & $+0.009$ \\
Infant states & 0.581 & 0.589 & $+0.008$ \\
Activity source  & 0.720 & 0.720 & $+0.000$ \\
\bottomrule
\end{tabular}}
\end{table}
\begin{table*}[t]
\centering
\caption{Macro-F$_1$ of simple physiological baselines against the proposed model on the
fixed subject-disjoint split. BeatGraph values are means over five seeds.}
\label{tab:baselines}
\small
\adjustbox{max width=\linewidth}{%
\begin{tabular}{llcccc}
\toprule
& & Sleep & Infant states & Activity source & Infant affect \\
Family & Model & (2-class) & (4-class) & (2-class) & (2-class) \\
\midrule
\multirow{3}{*}{Feature-based}
 & Mean heart rate only                & 0.644 & 0.300 & 0.524 & 0.416 \\
 & 14 HRV features, logistic regression & 0.741 & 0.434 & 0.528 & 0.523 \\
 & 14 HRV features, gradient-boosted trees & 0.723 & 0.466 & 0.571 & 0.517 \\
\midrule
\multirow{2}{*}{Reduced-input neural}
 & RR-only (waveform ablated)          & 0.724 & 0.442 & 0.558 & 0.521 \\
 & From-scratch supervised (no pretraining) & 0.782 & 0.492 & 0.616 & 0.603 \\
\midrule
Proposed
 & BeatGraph (self-supervised, $N{=}80$) & \textbf{0.923} & \textbf{0.589} & \textbf{0.745} & \textbf{0.731} \\
\bottomrule
\end{tabular}}
\end{table*}
\subsection{Effect of validation-set construction}
\label{app:lofo-selection}

We show two sets of results: one using no validation set and one using one infant as validation set. All results in this
appendix select on a held-out validation infant. Table~\ref{tab:lofo-selection}
quantifies the difference on the fold whose test infant matches our fixed split, and shows that
the two protocols can differ by small margin.

%

\section{BeatGraph Implementation Details}
\label{app:impl}

\subsection{Signal-quality filtering and pretraining window count}
\label{app:impl-quality}

Quality control is applied at the level of whole recordings rather than individual windows.
Each recording's cached peak file is audited, and recordings yielding fewer than
\textbf{five} detected peaks are removed from the pretraining manifests together with their
peak files. After this filter, the pretraining pool contains \textbf{403{,}320} non-overlapping $30$~s windows.

We deliberately apply \emph{no} per-window signal-quality index and no artifact rejection:
the recordings are naturalistic day-long sessions in the home, and windows containing
motion artifact, electrode noise, or missing contact are retained. This keeps the
pretraining distribution representative of deployment conditions, at the cost of a noisier
objective. 

\subsection{Window labelling: mixed labels and transitions}
\label{app:impl-windows}

Downstream windows are enumerated \emph{within} each annotated segment rather than across
the whole recording: for a segment spanning $[t_{\text{start}}, t_{\text{end}})$ with a
single coder label, we emit non-overlapping $30$~s windows starting at
$t_{\text{start}}$ and stopping before $t_{\text{end}}$. Consequently
\textbf{mixed-label windows cannot arise} --- every window lies entirely inside one labelled
segment and inherits that segment's label unambiguously. The cost of this construction is
that the residual shorter than one window at the end of each segment is discarded, as are
segments shorter than $30$~s in total.

For the reported activity-source model we apply \textbf{no transition filter}: all
non-overlapping windows within a qualifying segment are retained, including the
steady-state interior of long segments. 

\subsection{Class balancing}
\label{app:impl-balance}

We do not use any class balancing techniques. Metrics are macro-averaged so that rare classes (crying, quiet)
contribute equally to the reported score.

\subsection{Self-supervised projection and prediction heads}
\label{app:impl-heads}

The self-distillation branch uses BYOL-style heads on top of the pooled window embedding.
The \textbf{projector} is a two-layer MLP,
$\mathrm{Linear}(d \rightarrow d) \rightarrow \mathrm{LayerNorm}(d) \rightarrow
\mathrm{ReLU} \rightarrow \mathrm{Linear}(d \rightarrow 128)$ with $d = 128$, so both its
hidden and output widths are $128$. The \textbf{predictor}, applied only on the online
branch, has the same shape at the projector's output width:
$\mathrm{Linear}(128 \rightarrow 128) \rightarrow \mathrm{LayerNorm}(128) \rightarrow
\mathrm{ReLU} \rightarrow \mathrm{Linear}(128 \rightarrow 128)$. The target encoder and
target projector are exponential-moving-average copies of their online counterparts
(decay $0.99$) with gradients disabled, and both are discarded after pretraining. The
masked-prediction branch uses a separate single linear decoder
$\mathrm{Linear}(d \rightarrow d)$, also discarded after pretraining.

\subsection{Optimisation, batch size and hardware}
\label{app:impl-hardware}

Pretraining uses AdamW with learning rate $10^{-3}$, weight decay $10^{-4}$, batch size
$\mathbf{128}$, five warm-up epochs followed by cosine decay to a $10^{-5}$ floor, and
$100$ epochs over sampled windows per epoch. Downstream fine-tuning uses AdamW
with batch size $\mathbf{32}$, differential learning rates ($10^{-3}$ for the task head and
$5 \times 10^{-5}$ for the encoder and readout), weight decay $10^{-4}$, and $20$ epochs.

All runs use a \textbf{single GPU} --- an NVIDIA A100 (40~GB) or A40 --- with eight to
sixteen dataloader worker processes. One pretraining run takes approximately $22$--$23$
wall-clock hours; one downstream task-and-seed takes $15$--$20$ minutes, so a four-task,
five-seed sweep completes in about $1.5$ hours when the tasks run in parallel.

\section{Simple Physiological Baselines}
\label{app:physio-baselines}

A beat-graph encoder should be justified against the cardiac summaries a clinician or a
classical pipeline would compute first. We therefore compare against feature-based and
reduced-input baselines that isolate the two channels of information available in the
signal: interbeat-interval structure and beat morphology.

\subsection{Baseline definitions}
\label{app:baselines-defs}

\noindent\textbf{Mean heart rate.} A single scalar per window, the mean RR interval, passed to a
class-weighted logistic regression. This is the weakest physiologically meaningful
predictor and establishes how much of each task is explained by average rate alone.

\noindent\textbf{Standard HRV features.} Fourteen statistics computed per window from the interbeat
intervals: \texttt{mean\_rr}, \texttt{sdnn}, \texttt{rmssd}, \texttt{pnn50},
\texttt{min\_rr}, \texttt{max\_rr}, \texttt{hr\_range}, \texttt{sd1}, \texttt{sd2},
\texttt{s\_area}, \texttt{lf\_power}, \texttt{hf\_power}, \texttt{lf\_hf\_ratio} and
\texttt{total\_power}, covering time-domain, Poincar\'e and spectral descriptors. These are
fed to a class-weighted logistic regression and, separately, to gradient-boosted decision
trees.

\noindent\textbf{Gradient-boosted trees on HR and HRV features.} The same fourteen features with a
gradient-boosting classifier, which captures the non-linear interactions a linear model
cannot and is the strongest non-neural competitor.

\noindent\textbf{RR-only neural model.} The full architecture with the beat waveform ablated: the
convolutional beat descriptor is replaced by zeros so that each node carries only its
interbeat-interval features. Pretraining and fine-tuning are otherwise identical, isolating
what the graph can learn from rhythm alone.

\noindent\textbf{From-scratch supervised model.} The identical architecture trained directly on each
labelled task from random initialisation, with no self-supervised pretraining. This
isolates the contribution of pretraining from that of the architecture.

\subsection{Results}
\label{app:baselines-results}

Table~\ref{tab:baselines} reports macro-F$_1$ on the fixed subject-disjoint split. Feature
baselines are evaluated on the same windows as the neural models, up to a small number of
windows discarded when interval features could not be computed.

\section{Additional Ablation}
\subsection{Effect of BYOL loss}
\begin{table}[t]
\centering
\caption{Effect of the self-distillation loss weight $\lambda_{\mathrm{BYOL}}$ on
downstream macro-F1. Bold marks $\lambda_{\mathrm{BYOL}}=0.5$, the value used in
all other experiments.}
\label{tab:byol-weight}
\small
\adjustbox{max width=\linewidth}{%
\begin{tabular}{ccccc}
\toprule
$\lambda_{\mathrm{BYOL}}$ & Sleep & State & Source & Affect \\
\midrule
0.0 & $0.84_{\pm 0.05}$ & $0.42_{\pm 0.02}$ & $0.65_{\pm 0.03}$ & $0.61_{\pm 0.02}$ \\
0.2 & $0.91_{\pm 0.06}$ & $0.56_{\pm 0.02}$ & $0.69_{\pm 0.04}$ & $0.71_{\pm 0.04}$ \\
\textbf{0.5} & $\mathbf{0.92_{\pm 0.04}}$ & $\mathbf{0.59_{\pm 0.03}}$
     & $\mathbf{0.75_{\pm 0.02}}$ & $\mathbf{0.73_{\pm 0.02}}$ \\
1.0 & $0.85_{\pm 0.04}$ & $0.38_{\pm 0.03}$ & $0.70_{\pm 0.02}$ & $0.63_{\pm 0.03}$ \\
\bottomrule
\end{tabular}}
\end{table}
Table~\ref{tab:byol-weight} varies the weight of the self-distillation
regularizer. Setting $\lambda_{\mathrm{BYOL}}=0$ leaves masked
beat-embedding prediction as the only objective, and because its target
comes from a stop-gradient pass over the same weights, nothing in that
objective alone prevents the encoder from mapping every beat to a constant.
Removing it costs 0.08 to 0.17 macro-F1 across the four tasks. On infant
state the cost is larger still: $\lambda_{\mathrm{BYOL}}=0$ reaches 0.42,
below the 0.49 obtained by the same architecture trained from scratch
(Table~\ref{tab:baselines}), so on the task with the finest label
granularity a degenerate pretraining objective is worse than no pretraining
at all. Performance is non-monotonic in $\lambda_{\mathrm{BYOL}}$, peaking
at $0.5$ and falling at $1.0$ on every task. Infant state is again the most
affected and the only task falling below the $\lambda_{\mathrm{BYOL}}=0$
baseline, though with four folds and a $0.04$ gap we do not read that
inversion as more than suggestive. We use $\lambda_{\mathrm{BYOL}}=0.5$
throughout.


\end{document}